\pdfoutput=1

\documentclass[11pt]{article}

\usepackage[]{acl}

\usepackage{times}
\usepackage{latexsym}

\usepackage[T1]{fontenc}

\usepackage[utf8]{inputenc}
\usepackage{verbatim}
\usepackage{color}
\usepackage[usenames,dvipsnames]{xcolor}
\usepackage{tabularx}
\usepackage{tabu}
\usepackage{booktabs}
\usepackage{multirow}

\usepackage{graphicx} 
\usepackage{float}
\usepackage{subfigure} 
\usepackage{amssymb}

\usepackage{hyperref}
\usepackage{tablefootnote}
\usepackage{xspace}

\usepackage{threeparttable}
\usepackage{float}
\usepackage{amsmath}
\usepackage{bm}
\usepackage{algorithmicx}
\usepackage{algorithm}
\usepackage{algpseudocode}

\usepackage{arydshln}

\usepackage{amssymb}
\usepackage{pifont}
\newcommand{\xmark}{\ding{55}}%
\usepackage{enumitem}
\usepackage{tcolorbox}
\tcbuselibrary{skins}
\usepackage{placeins}
\usepackage{stfloats}
\usepackage{url}
\usepackage{microtype}

\newcommand{\ours}{\textsc{EDIR}\xspace}
\newcommand{\nquery}{5,000\xspace}
\newcommand{\ncorpus}{178,645\xspace}
\newcommand{\ourmodel}{\textsc{EDIR-MLLM}\xspace}

\newcommand{\nmodel}{13\xspace}
\newcommand{\cirr}{\textsc{CIRR}\xspace}
\newcommand{\circo}{\textsc{CIRCO}\xspace}
\newcommand{\fashioniq}{\textsc{FashionIQ}\xspace}
\newcommand{\genecis}{\textsc{GeneCIS}\xspace}
\newcommand{\icir}{\textsc{I-CIR}\xspace}
\newcommand{\magiclens}{\textsc{MagicLens}\xspace}
\newcommand{\picword}{\textsc{pic2word}\xspace}
\newcommand{\searle}{\textsc{searle}\xspace}

\newcommand{\cir}{\textsc{CIR}\xspace}
\newcommand{\eg}{\hbox{e.g.,}\xspace}
\newcommand{\ie}{\hbox{i.e.,}\xspace}

\title{Rethinking Composed Image Retrieval Evaluation: A Fine-Grained Benchmark from Image Editing}

\author{
\textbf{
Tingyu Song $^{\hspace{0.02em}{{\boldsymbol{1}}}}$$^{\hspace{0.02em}{{\boldsymbol{2}}}}$$^{\hspace{0.02em}{{\boldsymbol{3}}}}$ \quad 
Yanzhao Zhang $^{\hspace{0.02em}{{\boldsymbol{2}}}}$\quad
Mingxin Li $^{\hspace{0.02em}{{\boldsymbol{2}}}}$ \quad
Zhuoning Guo$^{\hspace{0.02em}{{\boldsymbol{4}}}}$$^{\hspace{0.02em}{{\boldsymbol{2}}}}$\quad
}\\
\textbf{
Dingkun Long $^{\hspace{0.02em}{{\boldsymbol{2}}}}$\quad 
Pengjun Xie $^{\hspace{0.02em}{{\boldsymbol{2}}}}$ \quad
Siyue Zhang $^{\hspace{0.02em}{{\boldsymbol5}}}$ \quad
Yilun Zhao$^{\hspace{0.02em}{{\boldsymbol{6}}}}$ \quad
Shu Wu$^{\hspace{0.03em}{{\boldsymbol{1}}}}$$^{\hspace{0.02em}{{\boldsymbol{3}}}}$
} 
\vspace{9pt}\\
$^{\hspace{0.02em}{{\boldsymbol{1}}}}$CASIA \quad
$^{\hspace{0.02em}{{\boldsymbol{2}}}}$Tongyi Lab, Alibaba Group \quad 
$^{\hspace{0.02em}{{\boldsymbol{3}}}}$UCAS \quad \\
$^{\hspace{0.02em}{{\boldsymbol{4}}}}$HKUST(GZ) \quad 
$^{\hspace{0.02em}{{\boldsymbol{5}}}}$NTU \quad
$^{\hspace{0.02em}{{\boldsymbol{6}}}}$Yale \quad
}

\begin{document}
\maketitle
\begin{abstract}
Composed Image Retrieval (\cir) is a pivotal and complex task in multimodal understanding. Current \cir benchmarks typically feature limited query categories and fail to capture the diverse requirements of real-world scenarios. To bridge this evaluation gap, we leverage image editing to achieve precise control over modification types and content, enabling a pipeline for synthesizing queries across a broad spectrum of categories. Using this pipeline, we construct \textbf{\ours}, a novel fine-grained \cir benchmark. \ours encompasses 5,000 high-quality queries structured across five main categories and fifteen subcategories. Our comprehensive evaluation of \nmodel multimodal embedding models reveals a significant capability gap; even state-of-the-art models (\eg RzenEmbed and GME) struggle to perform consistently across all subcategories, highlighting the rigorous nature of our benchmark. Through comparative analysis, we further uncover inherent limitations in existing benchmarks, such as modality biases and insufficient categorical coverage. Furthermore, an in-domain training experiment demonstrates the feasibility of our benchmark. This experiment clarifies the task challenges by distinguishing between categories that are solvable with targeted data and those that expose the intrinsic limitations of current model architectures\footnote{Correspondence: Shu Wu (\texttt{shu.wu@nlpr.ia.ac.cn} ). Code and Data will be available at \url{https://github.com/sighingsnow/edir}.} 
\end{abstract}

\section{Introduction}
\begin{figure*}[!t]
    \centering
    \includegraphics[width=\textwidth]{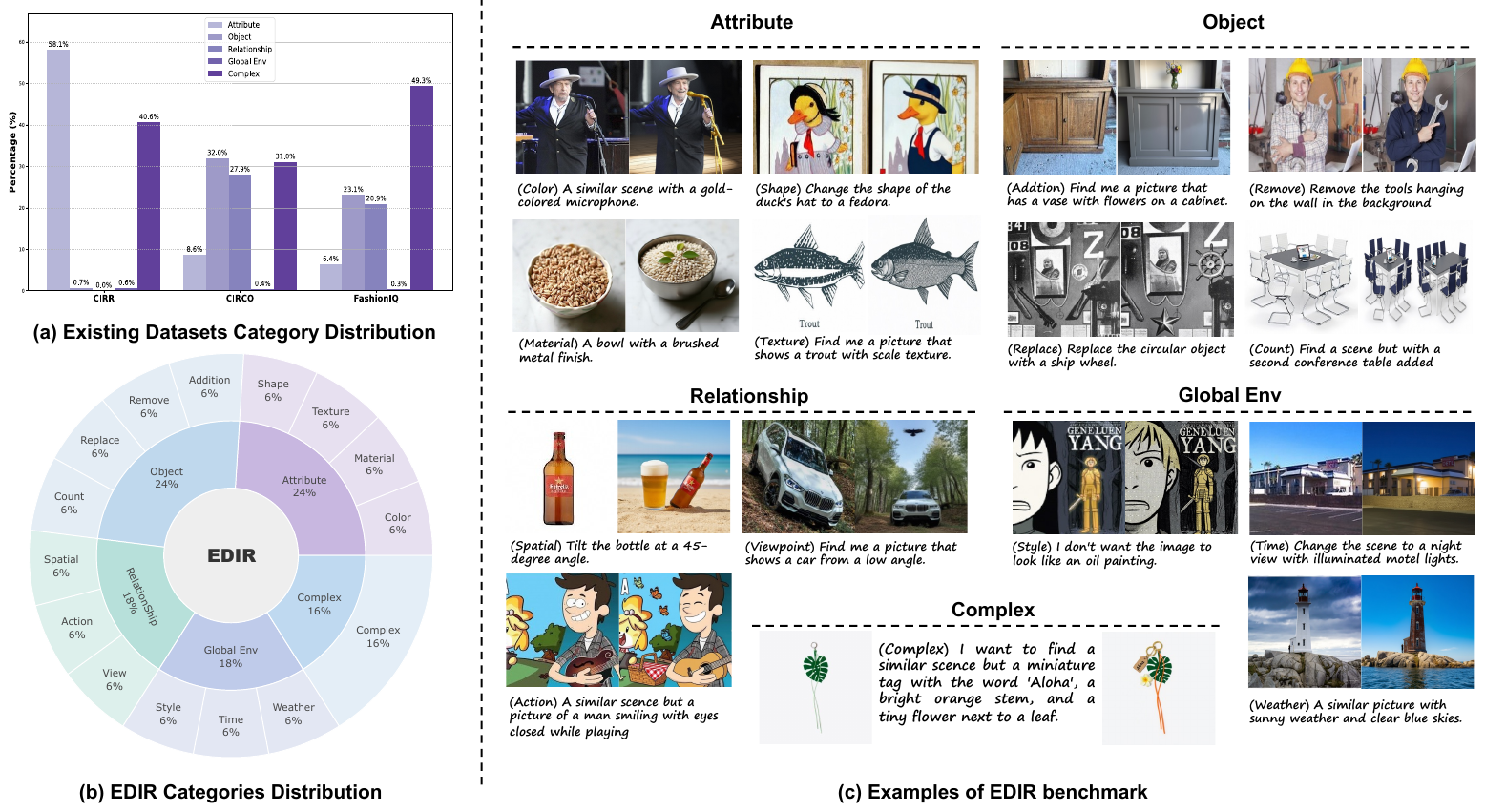}
    \caption{(a) Query category distribution in existing benchmarks, re-categorized using our taxonomy. (b) The balanced distribution of \ours across five main categories and fifteen subcategories. (c) Example queries illustrating the fine-grained nature of each subcategory. Left is the source image, right is the target image, and the below text is the \cir query text.  }
    \label{fig-overview}
\end{figure*}
Composed Image Retrieval (\cir) aims to retrieve a target image given a query composed of a reference image and a natural language description that specifies a desired modification~\cite{du2025survey, song2025comprehensive, wan2025composed}. This task has attracted increasing research interest due to its broad applicability in domains such as web search, interactive editing, and e-commerce. Consequently, a number of benchmarks have been proposed to evaluate \cir models, including CIRR~\cite{cirr}, FashionIQ~\cite{fashioniq}, and CIRCO~\cite{circo}. 

Despite these contributions, current \cir benchmarks suffer from two primary drawbacks. (1) \textbf{Coarse-grained Evaluation}: Existing benchmarks~\cite{cirr, fashioniq} provide a coarse-grained evaluation by focusing on a narrow range of modification categories. As a result, they neglect the broader spectrum of real-world requirements, as shown in \autoref{fig-overview}(a). (2) \textbf{Limited Query Scale}: While some benchmarks~\cite{circo, fashioniq} introduce query categories, they often suffer from insufficient scale and ambiguous category definitions. For instance, many queries in \circo are labeled with the \textit{``direct addressing''} tag, which often overlaps with more specific categories (\ie \textit{``color''}), thereby diluting the granularity of the evaluation. These limitations largely stem from the methodology used to construct these datasets. Specifically, the standard approach retrieves a target image for a source image first, and then annotates a query post-hoc to describe the difference. This dependence on the retriever's output leads to the absence of certain modification categories and an insufficient number of queries for others. 

To address these limitations, we first propose a comprehensive taxonomy for \emph{fine-grained} \cir evaluation, organizing real-world requirements into five main categories and fifteen subcategories, as illustrated in \autoref{fig-overview}(b). We then introduce a novel data synthesis pipeline that leverages image editing~\cite{instructpix2pix,qwenedit} to populate this taxonomy. By initiating the process with a textual modification to synthesize the target image, our pipeline provides precise control over query types and content. Using this pipeline, we construct \textbf{\ours}, an Image \textbf{\underline{E}}diting \textbf{\underline{D}}erived Benchmark for Composed \textbf{\underline{I}}mage \textbf{\underline{R}}etrieval, a comprehensive CIR benchmark comprising \nquery high-quality queries and an image gallery of \ncorpus images.

We conduct an extensive evaluation of current multimodal embedding models on \ours. Our assessment includes models trained on Multimodal Large Language Models (MLLM)~\cite{qwen2vl, qwen25vl} and CLIP~\cite{clip}. We observe that even the top-performing models cannot consistently perform well across all subcategories. We attribute these shortcomings to both the inherent limitations of the models and a scarcity of suitable training data. Additionally, we compare \ours with existing \cir benchmarks and conduct a thorough analysis of their characteristics. We conclude that current benchmarks suffer from significant evaluation gaps, including insufficient coverage of fine-grained categories and modality bias, which allows models to achieve high scores by over-relying on text. 

Furthermore, to analyze the unique challenges presented by \ours, we conduct an in-domain training experiment. We train a model, \textbf{\ourmodel}, on our synthesized data. The resulting performance on \ours facilitates a critical analysis, allowing us to differentiate between challenges that can be overcome with sufficient in-domain data and those that expose fundamental, intrinsic limitations of current model architectures. 

We summarize our contributions as follows: 
\begin{itemize} [leftmargin=*]
\itemsep0em
\item We propose a comprehensive taxonomy for \cir and a controllable data synthesis pipeline that leverages image editing to populate it. 
\item  We introduce \ours, a new fine-grained benchmark designed to facilitate comprehensive evaluation in the CIR domain. 
\item We analyze current models and existing \cir benchmarks using \ours, revealing significant gaps in model capabilities and inherent limitations in prior benchmarks. 
\item We conduct an in-domain training experiment that provides insights for future model development by distinguishing between data-solvable challenges and intrinsic model weaknesses. 
\end{itemize}



\begin{table}[!t]
\centering
\renewcommand{\arraystretch}{1.05}
\scalebox{0.72}{ 
\begin{threeparttable}
\begin{tabular}{lrrc}
\toprule[.1em]
\textbf{Benchmark} & \textbf{\# Qry} & \textbf{\# Corpus}  & \textbf{\# Category} \\

\midrule
\cirr \cite{cirr} & 4,148 & 2,315 &  - \\
\fashioniq \cite{fashioniq} & 6,016 & 15,536 &  - \\
\circo \cite{circo} & 1,000 & 123,403 & 9\\
\icir \cite{icir} & 1,813 & NA$^1$ & 7 \\
\noalign{\vskip 0.5ex}\hdashline\noalign{\vskip 0.5ex}
\ours (ours) & 5,000 & 178,645 &  15 \\
\bottomrule[.1em]
\end{tabular}
\begin{tablenotes}[flushleft]
\item[1]\icir uses an instance-level corpus and does not report exact count.
\end{tablenotes}
\end{threeparttable}
}
\caption{Comparing \ours with previous benchmarks.}
\label{table-dataset-compare}
\end{table}

\section{Related Works}

\subsection{Composed Image Retrieval Benchmarks}
\label{related-works-benchmark}

The evaluation of \cir models is currently constrained by a limited number of available benchmarks. 
Early \cir benchmarks are either domain-specific or coarse-grained in query categories. 
For instance, FashionIQ~\cite{fashioniq} is confined to the fashion domain, while CIRR~\cite{cirr} provides a broader domain but lacks fine-grained \cir queries to diagnose model failures. 
To address these limitations, \circo~\cite{circo} introduces more detailed categories and careful annotations. 
However, it suffers from an imbalanced distribution of queries across categories and ambiguous category definitions. 
More recent benchmarks further extend CIR evaluation toward more complex reasoning settings. 
GeneCIS~\cite{genecis} is proposed to evaluate a model's capacity to dynamically adapt its understanding of ``similarity'' based on a textual condition. 
\icir~\cite{icir} introduces instance-level retrieval in the \cir task. However, it only provides seven categories, which dilutes the evaluation's granularity. 
Furthermore, many existing benchmarks exhibit a significant modality bias~\cite{collm, icir}, where models can achieve high scores by over-relying on text-only signals, failing to test genuine multimodal compositionality. 
Our work addresses the aforementioned evaluation gaps by introducing a comprehensive benchmark with wide coverage and fine-grained query categories. 

\subsection{Composed Image Retrieval Methods} 
Methods for \cir have evolved from specialized attribute classifiers~\cite{ak2018learning,yang2020generative, hou2021learning} to approaches built upon VLMs~\cite{clip, blip, blip2}. Current literature generally categorizes \cir approaches into three main streams: (1) \textit{Text-Inversion}~\cite{pic2word, circo, lincir}, which maps the reference image to a textual token for text-image fusion; (2) \textit{Data Synthesis}~\cite{covr2, magiclens, vista, megapairs}, which leverages generative models to create large-scale \cir triplets for training (though recent works~\cite{vista, compodiff} use this for training data, they do not address the benchmark evaluation gaps we identify); and (3) \textit{Training-Free Methods}~\cite{cirevl, weimocir, ldre}, which leverage modular pipelines combining off-the-shelf vision and language models for tasks such as captioning and zero-shot reasoning. 

Recently, MLLMs~\cite{gemini2, qwen25vl, internvl3} have shown strong performance on a wide range of tasks. 
Consequently, universal multimodal embedding models~\cite{mmembed, vlm2vec} have been developed based on MLLM architectures.
These MLLM-based models achieve superior performance on existing CIR benchmarks. 
However, given the restricted benchmark design and limited evaluation coverage mentioned above, it remains unclear whether the observed performance gains reflect genuine compositional reasoning or merely the exploitation of benchmark biases. This gap highlights the need for a more comprehensive and diagnostic evaluation framework for CIR. 

\begin{figure*}[!t]
    \centering
    \includegraphics[width=\textwidth]{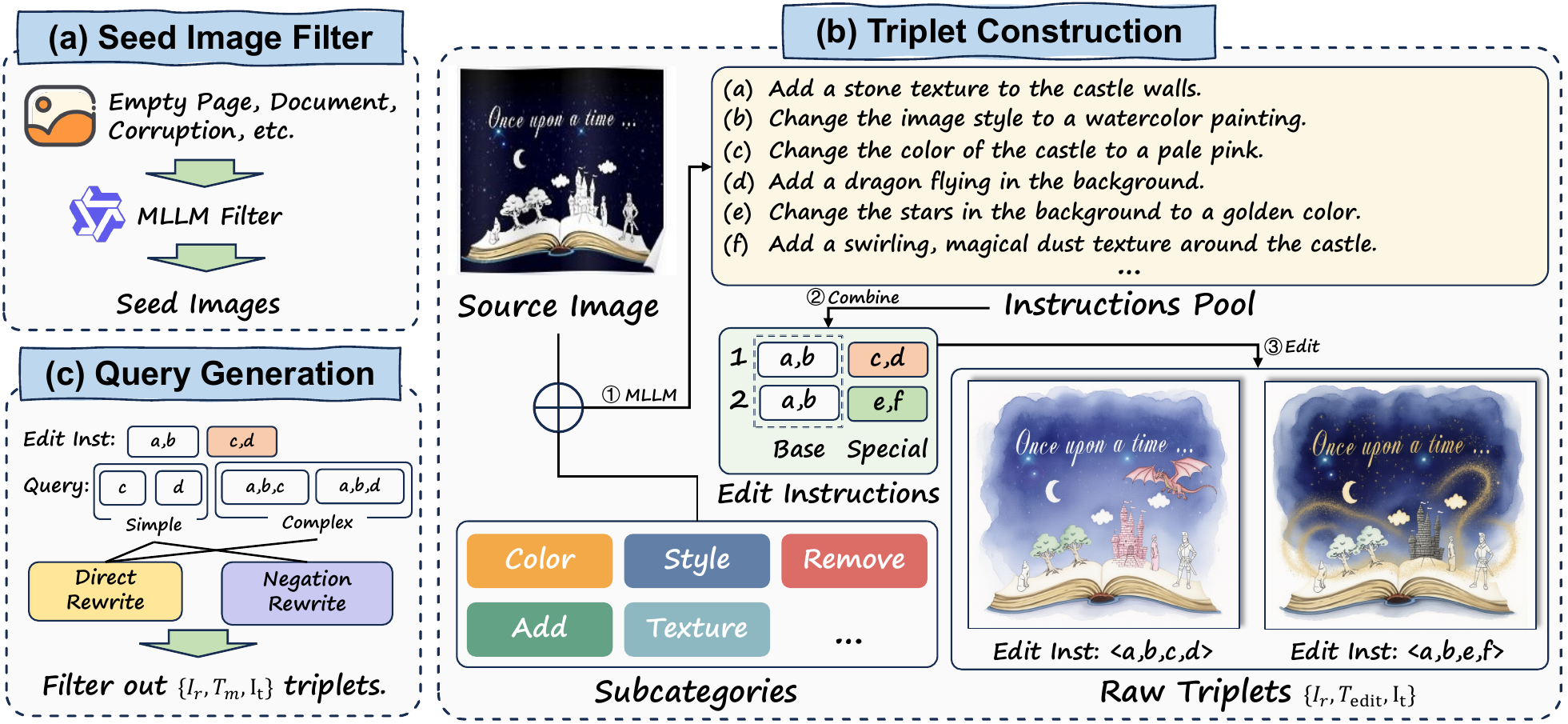}
    \caption{Overview of our data synthesis pipeline:  (1) Seed Image Selection: Unsuitable images are filtered from a large pool to select high-quality source images. (2) Triplet Generation: For each source image, multiple edit instructions are generated and applied to create ⟨source image, edit instruction, target image⟩ triplets. (3) Query Formulation: The edit instructions are automatically rewritten into natural language CIR queries.
    }
    \label{fig-data-synth-pipeline}
\end{figure*}

\section{\ours Benchmark Construction}

We propose \ours, a comprehensive benchmark designed to evaluate the capabilities of current multimodal embedding models in \cir in a fine-grained manner. 
Formally, each instance in our benchmark is a triplet $\{I_r, T_m, I_t\}$, comprising a reference image $I_r$, a target image $I_t$, and a text query $T_m$. 
Constructing such a benchmark at scale presents two primary technical challenges that are not adequately addressed by existing \cir benchmarks. 
First, we must ensure \emph{fine-grained, category-level diversity} to guarantee that the benchmark reflects the broad range of real-world \cir needs rather than a narrow set of edits. 
Second, we require a \emph{systematic and scalable} method to construct the text query $T_m$ for each category, ensuring that the modification is unambiguous, controllable, and aligned with the intended evaluation signal. 

To address these challenges, we first propose a comprehensive and hierarchical taxonomy that covers a wide spectrum of real-world modifications, as illustrated in \autoref{fig-overview}. 
Then, to systematically construct queries for these categories, we build an automated pipeline based on state-of-the-art image editing technology. 
This pipeline leverages our taxonomy to guide the generation process, ensuring that each resulting triplet is accurately aligned with a specific, fine-grained category. 
Specifically, we first generate an initial triplet $\{I_r, T_{edit}, I_t\}$, where a source image $I_r$ is modified to produce a target image $I_t$ based on a raw edit instruction $T_{edit}$ (\S\ref{subsec: cir-triplet}). 
Next, we refine the edit instruction $T_{edit}$ into a natural \cir query $T_m$(\S\ref{subsec: query-rewrite}). 
Finally, we apply a two-stage filtering process and human validation to ensure overall dataset quality (\S\ref{subsec: filter}, \S\ref{subsec: dataset-analysis}). 
 
\subsection{Preliminary Setup}
\paragraph{Taxonomy Definition.}  
Our taxonomy consists of five major categories: \textit{Attribute}, \textit{Object}, \textit{Relationship}, \textit{Global Environment}, and \textit{Complex}. 
(1) The \textit{Attribute} category focuses on object properties, mirroring e-commerce scenarios where a user might ask to see a product in a different color or material; 
(2) \textit{Object} involves operations such as adding or removing objects, which is fundamental for practical applications like photo editing and content creation; 
(3) \textit{Relationship} pertains to the spatial or semantic connections between objects, reflecting sophisticated needs like rearranging a scene for interior design or changing a viewpoint; 
(4) \textit{Global Environment} addresses holistic changes to the scene, such as style or weather, supporting creative searches for different moods or artistic effects; 
(5) \textit{Complex} queries combine multiple modifications from the other categories, representing the most realistic and challenging user requests that involve several simultaneous constraints. 
As detailed in \autoref{fig-overview} and \autoref{table-category-def}, these five categories are further broken down into fifteen subcategories. 

\paragraph{Seed Image Selection.} 
We select seed images from the LAION-400M~\cite{laion400m} dataset due to its vast coverage of real-world scenes. 
However, the dataset contains numerous corrupted, blank, or document-style images that are unsuitable for both CIR and image editing. 
To address this, we employ an MLLM (\ie Qwen2.5VL-32B) to automatically filter out these low-quality images. 
The specific prompt used for this filtering step is detailed in Appendix~\ref{appendix-prompts}. 

\begin{table}[!ht]
\centering
\small
\renewcommand{\arraystretch}{1.05}
\scalebox{0.78}{ 
\begin{tabular}{lp{7cm}} 
\toprule[.1em]
 \textbf{Subcategory} & \textbf{Definition} \\
\midrule
\multicolumn{2}{c}{\textbf{Category 1: Attribute}} \\
\noalign{\vskip 0.5ex}
(1)~~ Color & Changes the color of an object or an entire area. \\
(2)~~ Material & Modifies the surface material of an object. \\
(3)~~ Shape & Alters the geometric form or outline of an object. \\
(4)~~ Texture & Adds or alters fine surface details, patterns on an object. \\
\midrule
\multicolumn{2}{c}{\textbf{Category 2: Object}} \\
\noalign{\vskip 0.5ex}
(5)~~ Addition & Introduces a new, distinct object into the scene. \\
(6)~~ Remove & Completely eliminates an existing object or element. \\
(7)~~ Replace & Swaps an existing object with another object. \\
(8)~~ Count & Changes the number of instances of a specific object. \\
\midrule
\multicolumn{2}{c}{\textbf{Category 3: Relationship}} \\
\noalign{\vskip 0.5ex}
(9)~~ Spatial & Modifies the position, orientation, or background of elements,spatial relationships between items.  \\
(10) Action & Makes a subject in the image perform a new action. \\
(11) Viewpoint & Alters the camera's position, angle, or in-door and out-door transform. \\ 
\midrule
\multicolumn{2}{c}{\textbf{Category 4: Global Environment}} \\
\noalign{\vskip 0.5ex}
(12) Style & Changes the artistic style of the image. \\
(13) Time & Changes the time of day depicted in the scene.  \\
(14) Weather & Modifies the weather conditions shown in the image. \\
\midrule
\multicolumn{2}{c}{\textbf{Category 5: Complex}} \\
\noalign{\vskip 0.5ex}
(15) Complex & A query that combines two or more modifications from the simple categories above. \\
\bottomrule[.1em]
\end{tabular}
}
\caption{Detailed categories definitions. }
\label{table-category-def}
\end{table}

\subsection{Raw Triplet Construction}
\label{subsec: cir-triplet}

For each source image $I_r$, we first use an MLLM (\ie Qwen2.5-VL-32B) to identify 5-6 suitable subcategories from our taxonomy. For each suitable subcategory, the MLLM generates three distinct edit instructions, creating an instruction pool. Our core strategy leverages this pool to synthesize a set of related, complex images $\{I_1, I_2, ... I_n\}$ using an image editing model (\ie Qwen-Image-Edit~\cite{qwenedit}). Within this set, one image is randomly selected to serve as the \emph{target image} $I_t$ for a given query, while the others serve as \emph{hard negatives}. To achieve this, each image $I_i$ is generated by applying a composite of instructions $\{a,b,c,d\}$. One part consists of base modifications $\{a,b\}$ sampled from different categories; these establish a shared visual context crucial for our hard negative mining strategy. The second part consists of distinctive modifications $\{c,d\}$, which are randomly sampled to prevent the retrieval task from becoming trivial. If a target image were generated with only one unique change, the retrieval task would be overly simple for current \cir models. 
The full process is illustrated in~\autoref{fig-data-synth-pipeline}(b).

\subsection{Query Rewrite}
\label{subsec: query-rewrite}
Since the raw edit instructions are not directly suitable for use as CIR queries, we need to further refine the edit instruction into a natural \cir query. 
As established in \S\ref{subsec: cir-triplet}, each editing process is guided by a composite instruction $\{a,b,c,d\}$. The instructions $a$ and $b$ are basic operations that create a shared visual context, while $c$ and $d$ are the distinctive modifications. 
For simple queries, we use one of the distinctive modifications, $c$ or $d$, as the basis. For complex queries, we combine one distinctive modification with the two basic operations, resulting in a query based on $\{a,b,c\}$ or $\{a,b,d\}$, as illustrated in \autoref{fig-data-synth-pipeline}(c). We avoid using the full set of instructions $\{a,b,c,d\}$ for a single query, as this would make the query overly specific and could negatively impact retrieval performance. 
We utilize an LLM (\ie Qwen3-32B) to rewrite the edit instruction. 
Following previous work~\cite{megapairs,collm}, we employ several prompt templates to rephrase the edit instructions into natural language queries. 
In addition to direct rewrites, we recognize the importance of negation queries. For example, a positive query might be ``I want the same dress but in red,'' whereas a corresponding negation query could be ``Show me this dress in a different color.'' These types of queries are common in daily life, especially for categories such as \textit{Color} and \textit{Shape}. Therefore, we intentionally construct negation-based queries for these specific categories. 

\subsection{Data Quality Control}
\label{subsec: filter}
To improve data quality, we implement a two-stage filtering pipeline using an MLLM (\ie QwenVL-32B). 
The first stage occurs after the raw triplet construction (\autoref{fig-data-synth-pipeline}(b)). The MLLM assesses whether the generated image matches the full composite edit instruction, filtering out 312,009 of the 368,437 initial images. 
However, the complexity of these instructions occasionally allows partially correct images to pass. 
Therefore, a second filtering stage is applied after query rewriting (\autoref{fig-data-synth-pipeline}(c)).  
In this step, the MLLM re-evaluates the $\{I_r, I_t\}$ pair against the more concise \cir query $T_m$.  
This second pass filtered out 889,013 from 1,087,710 triplets, improving the final dataset's alignment with the queries. 
We provide details of the construction process in Appendix~\ref{appendix-edir-construction} and prompts in Appendix~\ref{appendix-prompts}. 

\begin{table*}[htbp]
\centering
\small
\footnotesize
\renewcommand{\arraystretch}{1.1}
\setlength{\tabcolsep}{2pt}
\scalebox{0.88}{
\begin{tabular}{lcccccccccccccccc}
\toprule[.1em]
\multirow{2}{*}{\textbf{Metric}} & \multirow{2}{*}{\textbf{Total}} & \multicolumn{4}{c}{\textbf{Attribute}} & \multicolumn{4}{c}{\textbf{Object}} & \multicolumn{3}{c}{\textbf{Relationship}} &  \multicolumn{3}{c}{\textbf{Style}} & \textbf{Complex} \\
 \cmidrule(lr){3-6}  \cmidrule(lr){7-10} \cmidrule(lr){11-13} \cmidrule(lr){14-16} \cmidrule(lr){17-17} 
 & & Color & Material & Shape & Texture & Add & Remove & Replace & Count & Spatial & Action & View & Style & Weather & Time & Complex \\
 
\midrule
\multicolumn{16}{c}{\emph{\textbf{Non MLLM-based Models}}} \\
\midrule
\picword & \textbf{21.2} & \textbf{22.0} & \textbf{15.3} & 18.0 & 16.7 & \textbf{28.7} & 11.7 & 24.3 & \textbf{23.3} & \textbf{19.0} & 27.7 & 12.0 & 21.3 & \textbf{26.7} & \textbf{29.3} & \textbf{21.8} \\
\searle & 17.1 & 20.7 & 15.0 & 15.7 & 12.0 & 25.0 & \textbf{8.3} & 21.3 & 20.0 & 10.7 & \textbf{28.7} & \textbf{8.0} & 10.0 & 22.0 & 20.0 & 18.1 \\
\magiclens & 16.8 & 19.3 & 10.3 & \textbf{9.7} & \textbf{6.7} & 22.0 & 12.0 & \textbf{29.3} & 18.3 & 18.0 & 24.3 & 10.0 & \textbf{23.3} & 11.0 & 18.7 & 17.4 \\
\noalign{\vskip 0.5ex}\hdashline\noalign{\vskip 0.5ex}
Avg. & 18.4 & 20.7 & 13.6 & 14.4 & 11.8 & 25.2 & 10.7 & 25.0 & 20.6 & 15.9 & 26.9 & 10.0 & 18.2 & 19.9 & 22.7 & 19.1 \\
\midrule
\multicolumn{16}{c}{\emph{\textbf{MLLM-based Models}}} \\
\midrule
RzenEmbed-7B & \textbf{47.2} & 44.7 & 37.3 & 35.7 & 36.7 & 74.0 & \textbf{28.0} & \textbf{71.0} & 49.0 & \textbf{45.7} & \textbf{60.7} & 24.0 & 44.3 & \textbf{50.7} & \textbf{58.0} & 47.5 \\
Ops-embedding & \textbf{47.2} & \textbf{45.7} & \textbf{38.3} & \textbf{38.7} & \textbf{40.0} & \textbf{75.0} & 23.3 & 66.3 & \textbf{50.3} & 44.0 & 59.7 & \textbf{24.7} & \textbf{46.3} & \textbf{50.7} & 52.3 & \textbf{49.0} \\
GME-2B & 42.4 & 39.0 & 35.3 & 34.3 & 37.3 & 65.0 & 21.7 & 63.3 & 47.3 & 34.0 & 56.3 & 22.0 & 42.0 & 44.7 & 50.3 & 42.8 \\
GME-7B & 40.1 & 36.7 & 34.3 & 30.7 & 29.7 & 63.3 & 23.3 & 62.7 & 45.3 & 36.7 & 52.0 & 24.3 & 35.7 & 42.0 & 48.0 & 39.0 \\
MMRet-MLLM & 36.8 & 36.3 & 26.3 & 26.7 & 27.3 & 57.7 & 18.7 & 54.3 & 40.3 & 36.7 & 40.7 & 20.7 & 23.0 & 42.7 & 45.0 & 43.6 \\
E5-V & 34.0 & 26.3 & 27.3 & 26.0 & 27.3 & 55.0 & 14.7 & 51.0 & 39.3 & 37.0 & 49.0 & 11.7 & 35.7 & 34.3 & 38.7 & 34.9 \\
VLM2Vec-2B & 32.4 & 32.7 & 25.0 & 26.7 & 32.3 & 55.7 & 18.7 & 33.7 & 36.0 & 32.7 & 35.3 & 18.0 & 23.7 & 34.7 & 38.7 & 36.0 \\
UniME-7B & 31.8 & 23.7 & 16.0 & 23.3 & 20.3 & 48.7 & 17.0 & 46.0 & 41.0 & 35.3 & 46.3 & 17.7 & 29.0 & 27.7 & 35.7 & 38.5 \\
UniME-2B & 28.6 & 28.0 & 23.0 & 21.0 & 21.0 & 44.3 & 17.3 & 49.7 & 33.7 & 22.7 & 40.7 & 14.7 & 21.7 & 23.3 & 30.3 & 31.8 \\
mmE5 & 28.1 & 20.7 & 21.0 & 23.0 & 24.7 & 41.0 & 19.3 & 39.7 & 26.3 & 28.7 & 36.7 & 20.3 & 30.3 & 31.3 & 31.0 & 28.1 \\
\noalign{\vskip 0.5ex}\hdashline\noalign{\vskip 0.5ex}
Avg. & 36.9 & 33.4 & 28.4 & 28.6 & 29.7 & 58.0 & 20.2 & 53.8 & 40.9 & 35.3 & 47.7 & 19.8 & 33.2 & 38.2 & 42.8 & 39.1 \\
\ourmodel & 59.9 & 57.7 & 59.0 & 44.0 & 56.3 & 86.0 & 37.7 & 74.3 & 58.3 & 48.0 & 71.0 & 33.0 & 66.7 & 76.3 & 72.0 & 59.1 \\
\bottomrule[.1em]
\end{tabular}
}
\caption{Recall@1 performance of models on \ours. \textit{Avg.} is computed as the average performance across categories for each type of models, excluding \ourmodel. 
} 
\label{table-main-results}
\end{table*}

\subsection{Dataset Analysis}
\label{subsec: dataset-analysis}
\paragraph{Statistics. }
We initially sample 70,000 images and generate 368,437 edited images. 
After filtering, 889,013 high-quality $\{I_r, T_m, I_t\}$ triplets are obtained. From these, we construct our benchmark by randomly sampling 300 queries for each of the 14 simple categories and 800 queries for the \textit{Complex} category, resulting in a total of 5,000 queries. 
For each query, we include its target image along with three hard negatives generated from the same source image. 
To ensure corpus diversity, this set is augmented with 150,000 additional edited images which are also derived from the 70,000 source images. 
The final benchmark comprises 5,000 queries and a corpus of 178,645 images. 

\paragraph{Human Validation. } 
To assess dataset quality, we conduct a human validation study on a randomly selected 12\% sample. In this study, annotators evaluate three primary error types. The \emph{False Positive Rate} measures instances where the target image $I_t$ does not match the query $\{I_r, T_m\}$. 
The \emph{False Negative Rate} identifies cases where a provided hard negative image also satisfies the query. 
Finally, to measure the \emph{Global False Negative Rate}, we use a state-of-the-art \cir model~\cite{megapairs}, MMRet-MLLM, to retrieve the top-5 images from the corpus for each query $\{I_r, T_m\}$. Annotators then check if any of these retrieved images, other than the target image $I_t$, are also positive. 
Our manual annotation reveals a False Positive Rate of 8.0\%, a False Hard Negative Rate of 7.3\%, and a Global False Negative Rate of 11.7\%.
\section{Experiments and Results}

\subsection{Experiment Setup} 
We evaluate a wide range of multimodal models using Recall@1, including both MLLM-based and Non-MLLM-based types, as follows:
\paragraph{Non-MLLM-based Models.} We evaluate the following Non-MLLM-based methods and models: (1) \textbf{\picword}~\cite{pic2word}, which implements the \textit{text-inversion method} for \cir. (2) \textbf{SEARLE}~\cite{circo}, which is also based on the \textit{text-inversion method}. (3) \textbf{\magiclens}~\cite{magiclens}, which is trained on a large scale of \cir triplets.

\paragraph{MLLM-based Models.} We evaluate the following frontier MLLM-based models: (1) \textbf{GME-Qwen2-VL}~\cite{gme}, for which we include both the 2B and 7B versions. (2) \textbf{BGE-VL}~\cite{megapairs}, where we use the BGE-VL-MLLM-S1 version, which is not further finetuned on MMEB. (3) \textbf{VLM2Vec}~\cite{vlm2vec}, where we use VLM2Vec-V2.0, which is based on Qwen2-VL-2B. (4) \textbf{Ops-embedding}~\cite{ops}, where we use Ops-MM-embedding-v1-7B for evaluation, which shows competitive performance on relevant tasks. (5) \textbf{E5-V}~\cite{e5v}, where we use the 7B version of E5-V. (6) \textbf{UniME}~\cite{unimev2}, where we use UniME-Qwen2-VL and include both the 2B and 7B versions. (7) \textbf{mmE5}~\cite{mme5}, where we use the model trained based on Llama-3.2-11B-Vision. (8) \textbf{RzenEmbed-7B}~\cite{rzenembed}, where we use the RzenEmbed-V2-7B based on Qwen2-VL. 

\subsection{Results}

\paragraph{Non-MLLM-based Models. } 
Non-MLLM-based models achieve an average total score of only 18.4\%. We attribute this underperformance primarily to the limitations of the CLIP architecture upon which these models are built. Since many candidate images in \ours are visually similar edits of a single source, these models can identify the correct group of images but cannot accurately distinguish the target based on the fine-grained text query. This fundamental limitation explains their low scores in nuanced categories like \textit{remove} and \textit{texture}. This confirms that \ours is also a challenging benchmark for Non-MLLM-based models. 

\paragraph{MLLM-based Models.}  From \autoref{table-main-results}, we observe that MLLM-based models consistently outperform Non-MLLM baselines. They achieve relatively strong performance on the \textit{addition}, \textit{replace}, and \textit{action} categories. However, they perform notably worse on others, especially \textit{texture}, \textit{remove}, and \textit{shape}. We therefore conduct a detailed error analysis of these models (\S\ref{model-analysis}). In addition, to verify that \ours is a meaningful and complementary benchmark, we compare model performance on \ours against existing CIR benchmarks and provide a thorough analysis (\S\ref{benchmark-analysis}). 

\subsection{Error Analysis}
\label{model-analysis}
To better understand the current weaknesses of multimodal embedding models, we examine cases with low Recall@1 scores and develop the following taxonomy of error types. 
(1) \textbf{Failure in Handling Negation}: Models consistently struggle with queries involving negation, both in removal commands (e.g., ``remove the hat'') and with explicit negative terms (e.g., ``not red'').
(2) \textbf{Deficiencies in Compositional Reasoning}: Models exhibit poor performance on categories like \textit{count}, \textit{spatial}, \textit{style}, and \textit{viewpoint}. These tasks demand a form of compositional reasoning. The model must correctly interpret relationships between objects (\textit{spatial}, \textit{count}) or apply global transformations that affect the entire scene (\textit{style}, \textit{viewpoint}). For instance, executing a \textit{viewpoint} query to change an indoor scene to an outdoor one requires the model to reason about the global scene context and its constituent elements. This is a capability that current models appear to lack. 
(3) \textbf{Struggles with Multiple Constraints}: In the \textit{complex} category, queries provide multiple conditions. Models often retrieve images that only partially satisfy all constraints. This indicates a weakness in composing and verifying multiple distinct instructions from a single query. 
(4) \textbf{Insensitivity to Fine-Grained Details}: For categories such as \textit{texture}, \textit{material}, and \textit{shape}, the distinctions between the source and target images can be subtle. Current models tend to overlook these fine-grained visual changes, leading to errors. We provide a detailed error case study in the Appendix~\ref{appendix-error-analysis}. 

This underperformance stems from two interconnected issues: intrinsic model weaknesses and inadequate training data. Weaknesses in the foundational MLLMs~\cite{mme, videomme} explain the observed \textbf{Failure in Handling Negation} and \textbf{Deficiencies in Compositional Reasoning}, as these base models inherently struggle with logical and spatial operations. Simultaneously, the embedding models' inability to handle \textbf{Multiple Constraints} and \textbf{Fine-Grained Details} is exacerbated by training on data that lacks such complexity. This highlights the critical need for more carefully curated datasets to address these specific shortcomings and enhance model capabilities.

\subsection{Benchmark Analysis}
\label{benchmark-analysis}
\begin{figure}[!t]
    \centering
    \small
    \includegraphics[width=0.43\textwidth]{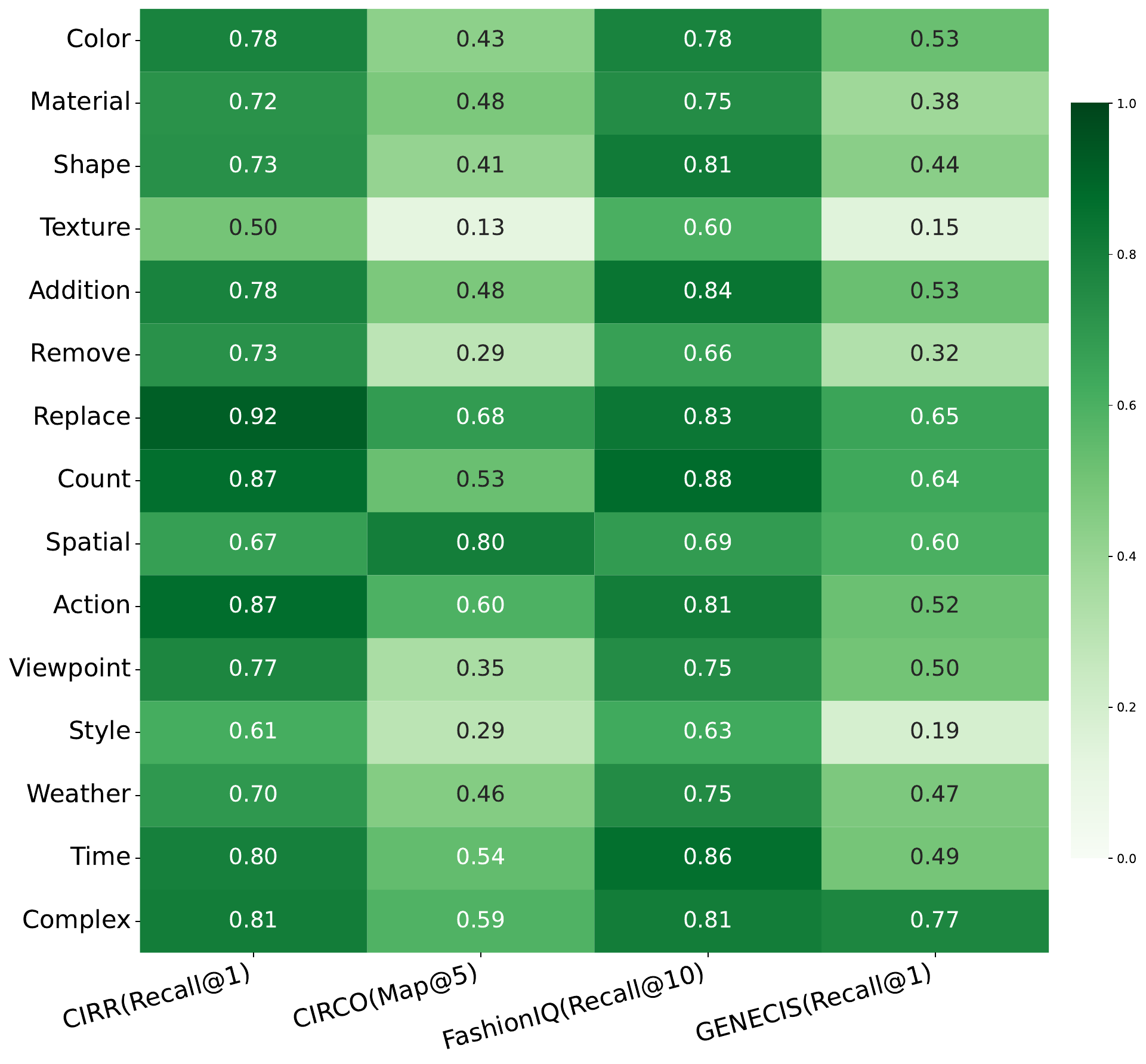}
    \caption{Performance correlation of MLLM-based models between \ours and prior CIR benchmarks.
} 
    \label{fig-corr-analysis}
\end{figure}

 To better understand the limitations of existing CIR benchmarks, we analyze the performance correlation of MLLM-based models across \ours and four prominent CIR benchmarks: \circo, \cirr, \fashioniq, and \genecis. 
 We compute the Spearman correlation coefficients between model performances. Performance on each benchmark is measured using its respective standard metric, including Recall@1 for \ours. 
 For \cirr and \circo, we use their validation sets to measure performance. As shown in ~\autoref{fig-corr-analysis}, \ours has a positive value between all categories and the target models. This verifies that \ours is qualified to evaluate the \cir abilities of current models. However, the results also reveal varying correlations. This confirms the two critical limitations of existing benchmarks mentioned in \S\ref{related-works-benchmark}: a fine-grained evaluation bias and a significant modality bias.

\paragraph{Fine-grained Evaluation Bias.}
Existing benchmarks lack balanced, fine-grained evaluation. Using an LLM (i.e., Qwen-32B) to classify their queries, we find a heavy skew towards \textit{complex} modifications, as shown in \autoref{fig-overview}. Meanwhile, they lack sufficient coverage of specific categories like \textit{remove}, \textit{spatial}, and \textit{texture}. For example, \circo only has 10 \textit{remove} queries, and \cirr has no \textit{spatial} queries in its validation set. This overall categorical imbalance helps explain why, in our correlation analysis illustrated in ~\autoref{fig-corr-analysis}, the performance correlation for these specific abilities is consistently lower relative to other categories within the same benchmark's results. This indicates that \ours addresses a critical evaluation gap by providing comprehensive coverage of these overlooked compositional skills. 

\paragraph{Modality Bias.} 
\begin{figure}[!t]
    \centering
    \small
    \includegraphics[width=0.32\textwidth]{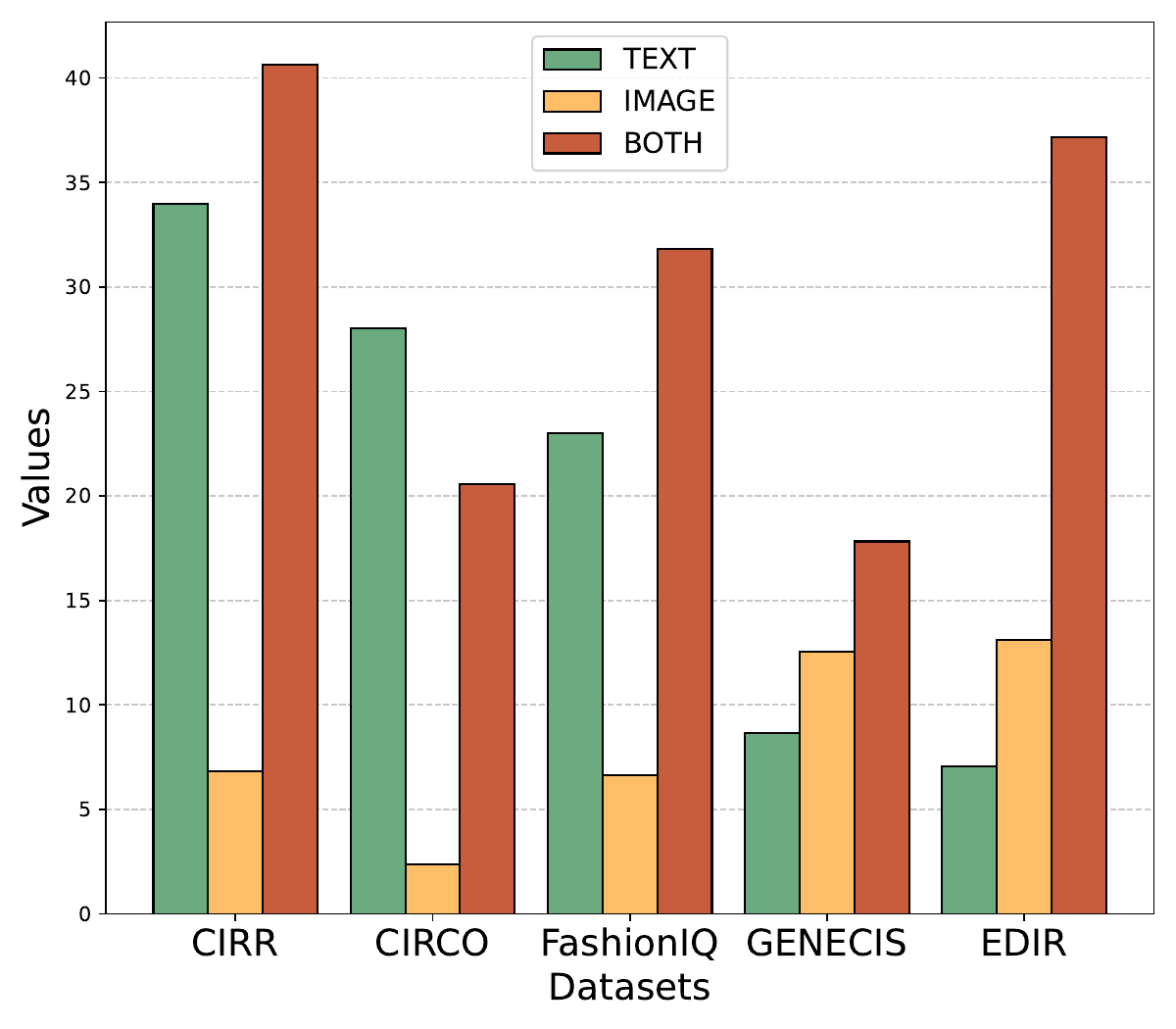}
    \caption{
    Average performance of MLLM-based models across CIR benchmarks. }
    \label{fig-modality-bias}
\end{figure}
Existing benchmarks can also exhibit a strong modality bias. We test this by evaluating MLLM-based models in \textit{text-only}, \textit{image-only}, and \textit{text-image} modes. As illustrated in \autoref{fig-modality-bias}, on \circo, models perform even better with only text, indicating the reference image is almost redundant. This text-centric shortcut also partially explains \circo's low correlation with \ours, which requires a genuine synthesis of both modalities and thus offers a more robust test of the \cir task. In conclusion, \ours provides a more fine-grained evaluation that simultaneously demands a compositional understanding of both image and text. 
\section{In-domain Training and Analysis}

To further investigate the unique challenges posed by \ours and its relationship with existing benchmarks, we conduct an in-domain training experiment. This experiment is designed to assess the solvability of \ours's fine-grained categories when a model is trained on specialized data. Leveraging our data synthesis pipeline~\autoref{fig-data-synth-pipeline}, we generated an additional pool of approximately 1.1 million high-quality edit triplets. From this pool, we curate a specialized training set by sampling 15,000 triplets for each of our 15 categories, totaling 225,000 training instances. We train a model based on Qwen2.5-VL~\cite{qwen25vl}, which we refer to as \ourmodel, on this dataset for 2,500 steps with a batch size of 128. We provide training details in Appendix~\ref{appendix-training-settings}. 

To determine if the challenges in \ours are solvable and to identify which categories remain difficult, we define a category as \emph{solvable} if its Recall@1 exceeds 60\% or shows an improvement of over 20 percentage points after in-domain training. The in-domain performance of \ourmodel demonstrates that our benchmark is indeed solvable. As shown in ~\autoref{table-main-results}, \ourmodel achieves a new state-of-the-art Recall@1 of 59.9\% on \ours. This is a substantial improvement over the average of other MLLM-based methods, which is 36.9\%. To gain a more granular understanding of these results, we analyze the performance on a per-category basis. These results directly corroborate our model analysis in \S\ref{model-analysis}. As mentioned, categories requiring sensitivity to fine-grained details, such as \textit{color}, \textit{material}, \textit{texture}, and \textit{action}, see dramatic improvements. This confirms our hypothesis that such challenges, often stemming from inadequate training data, can be largely overcome with corresponding examples. Conversely, categories demanding complex compositional reasoning, including \textit{count}, \textit{spatial}, and \textit{viewpoint}, exhibit modest gains. These issues represent intrinsic model weaknesses in operations involving reasoning, which are not easily resolved even with in-domain data. This illustrates that \ours can effectively distinguish between data-solvable challenges and the more fundamental architectural limitations of current models. 
\section{Conclusion}

We introduce \ours, a large-scale benchmark specifically designed for the granular evaluation of Composed Image Retrieval (\cir) tasks. 
Constructed through an innovative automated data synthesis pipeline that leverages image editing, \ours comprises \nquery queries across fifteen detailed subcategories. 
Our comprehensive evaluation of \nmodel multimodal embedding models reveals their significant shortcomings on \ours, highlighting a clear gap in current model capabilities regarding compositional generalization. Furthermore, a thorough comparison against existing \cir benchmarks confirms that \ours effectively uncovers model weaknesses that other evaluations overlooked. 
Finally, to validate the unique challenges posed by \ours, we conduct an in-domain training experiment. This not only demonstrates the solvability of \ours but also reveals its ability to distinguish between data-solvable issues and intrinsic model limitations. 
In conclusion, \ours provides the community with a robust tool to drive the development of more genuinely compositional and less biased \cir models. 
\section*{Limitations}

While our work introduces a fine-grained benchmark for Composed Image Retrieval (CIR), we acknowledge several limitations that open avenues for future research. 
First, a key limitation is the cost and scalability of our data synthesis pipeline. Although leveraging programmatic image editing provides precise control over modifications, the process remains computationally expensive, making large-scale data generation a challenge. 
Second, the complexity of our \textit{Complex} queries is bounded. The queries in our \ours benchmark are typically composed of three distinct conditions. While more challenging than single-edit queries, they do not yet represent highly complex scenarios with four or more interdependent instructions. This presents an opportunity to develop even more challenging benchmarks. 
Finally, our work is intentionally focused on evaluation. We designed \ours primarily as a benchmark to diagnose model weaknesses, rather than as a universal training solution. The development of scalable training methods tailored to address these weaknesses remains an open research direction. 
In conclusion, while our benchmark serves as an important diagnostic tool, addressing these limitations in scalability, complexity, and training will be crucial for advancing the next generation of CIR models. 

\bibliography{custom}

\begin{thebibliography}{43}
\expandafter\ifx\csname natexlab\endcsname\relax\def\natexlab#1{#1}\fi

\bibitem[{Ak et~al.(2018)Ak, Kassim, Lim, and Tham}]{ak2018learning}
Kenan~E Ak, Ashraf~A Kassim, Joo~Hwee Lim, and Jo~Yew Tham. 2018.
\newblock Learning attribute representations with localization for flexible fashion search.
\newblock In \emph{Proceedings of the IEEE conference on computer vision and pattern recognition}, pages 7708--7717.

\bibitem[{Bai et~al.(2025)Bai, Chen, Liu, Wang, Ge, Song, Dang, Wang, Wang, Tang et~al.}]{qwen25vl}
Shuai Bai, Keqin Chen, Xuejing Liu, Jialin Wang, Wenbin Ge, Sibo Song, Kai Dang, Peng Wang, Shijie Wang, Jun Tang, et~al. 2025.
\newblock Qwen2. 5-vl technical report.
\newblock \emph{arXiv preprint arXiv:2502.13923}.

\bibitem[{Baldrati et~al.(2023)Baldrati, Agnolucci, Bertini, and Del~Bimbo}]{circo}
Alberto Baldrati, Lorenzo Agnolucci, Marco Bertini, and Alberto Del~Bimbo. 2023.
\newblock Zero-shot composed image retrieval with textual inversion.
\newblock In \emph{Proceedings of the IEEE/CVF International Conference on Computer Vision}, pages 15338--15347.

\bibitem[{Brooks et~al.(2023)Brooks, Holynski, and Efros}]{instructpix2pix}
Tim Brooks, Aleksander Holynski, and Alexei~A Efros. 2023.
\newblock Instructpix2pix: Learning to follow image editing instructions.
\newblock In \emph{Proceedings of the IEEE/CVF conference on computer vision and pattern recognition}, pages 18392--18402.

\bibitem[{Chen et~al.(2025)Chen, Wang, Yang, Zhu, Zhao, Wei, and Dou}]{mme5}
Haonan Chen, Liang Wang, Nan Yang, Yutao Zhu, Ziliang Zhao, Furu Wei, and Zhicheng Dou. 2025.
\newblock mme5: Improving multimodal multilingual embeddings via high-quality synthetic data.
\newblock \emph{arXiv preprint arXiv:2502.08468}.

\bibitem[{Du et~al.(2025)Du, Deng, Li, Li, and Tian}]{du2025survey}
Longye Du, Shuaiyu Deng, Ying Li, Jun Li, and Qi~Tian. 2025.
\newblock A survey on composed image retrieval.
\newblock \emph{ACM Transactions on Multimedia Computing, Communications and Applications}.

\bibitem[{Fu et~al.(2025{\natexlab{a}})Fu, Chen, Shen, Qin, Zhang, Lin, Yang, Zheng, Li, Sun et~al.}]{mme}
Chaoyou Fu, Peixian Chen, Yunhang Shen, Yulei Qin, Mengdan Zhang, Xu~Lin, Jinrui Yang, Xiawu Zheng, Ke~Li, Xing Sun, et~al. 2025{\natexlab{a}}.
\newblock Mme: A comprehensive evaluation benchmark for multimodal large language models.
\newblock In \emph{The Thirty-ninth Annual Conference on Neural Information Processing Systems Datasets and Benchmarks Track}.

\bibitem[{Fu et~al.(2025{\natexlab{b}})Fu, Dai, Luo, Li, Ren, Zhang, Wang, Zhou, Shen, Zhang et~al.}]{videomme}
Chaoyou Fu, Yuhan Dai, Yongdong Luo, Lei Li, Shuhuai Ren, Renrui Zhang, Zihan Wang, Chenyu Zhou, Yunhang Shen, Mengdan Zhang, et~al. 2025{\natexlab{b}}.
\newblock Video-mme: The first-ever comprehensive evaluation benchmark of multi-modal llms in video analysis.
\newblock In \emph{Proceedings of the Computer Vision and Pattern Recognition Conference}, pages 24108--24118.

\bibitem[{Google(2024)}]{gemini2}
Google. 2024.
\newblock \href {https://blog.google/technology/google-deepmind/google-gemini-ai-update-december-2024/} {Gemini-2.0}.

\bibitem[{Gu et~al.(2024{\natexlab{a}})Gu, Chun, Kim, , Kang, and Yun}]{lincir}
Geonmo Gu, Sanghyuk Chun, Wonjae Kim, , Yoohoon Kang, and Sangdoo Yun. 2024{\natexlab{a}}.
\newblock Language-only training of zero-shot composed image retrieval.
\newblock In \emph{Conference on Computer Vision and Pattern Recognition (CVPR)}.

\bibitem[{Gu et~al.(2024{\natexlab{b}})Gu, Chun, Kim, Jun, Kang, and Yun}]{compodiff}
Geonmo Gu, Sanghyuk Chun, Wonjae Kim, HeeJae Jun, Yoohoon Kang, and Sangdoo Yun. 2024{\natexlab{b}}.
\newblock \href {https://openreview.net/forum?id=mKtlzW0bWc} {Compodiff: Versatile composed image retrieval with latent diffusion}.
\newblock \emph{Transactions on Machine Learning Research}.
\newblock Expert Certification.

\bibitem[{Gu et~al.(2025)Gu, Yang, Zhang, An, Feng, Zhang, Cai, Deng, and Bing}]{unimev2}
Tiancheng Gu, Kaicheng Yang, Kaichen Zhang, Xiang An, Ziyong Feng, Yueyi Zhang, Weidong Cai, Jiankang Deng, and Lidong Bing. 2025.
\newblock Unime-v2: Mllm-as-a-judge for universal multimodal embedding learning.
\newblock \emph{arXiv preprint arXiv:2510.13515}.

\bibitem[{Hafner et~al.(2021)Hafner, Katsantoni, K{\"o}ster, Marks, Mukherjee, Staiger, Ule, and Zavolan}]{clip}
Markus Hafner, Maria Katsantoni, Tino K{\"o}ster, James Marks, Joyita Mukherjee, Dorothee Staiger, Jernej Ule, and Mihaela Zavolan. 2021.
\newblock Clip and complementary methods.
\newblock \emph{Nature Reviews Methods Primers}, 1(1):20.

\bibitem[{Hou et~al.(2021)Hou, Vig, Donoser, and Bazzani}]{hou2021learning}
Yuxin Hou, Eleonora Vig, Michael Donoser, and Loris Bazzani. 2021.
\newblock Learning attribute-driven disentangled representations for interactive fashion retrieval.
\newblock In \emph{Proceedings of the IEEE/CVF International conference on computer vision}, pages 12147--12157.

\bibitem[{Huynh et~al.(2025)Huynh, Yang, Tawari, Shah, Tran, Hamid, Chilimbi, and Shrivastava}]{collm}
Chuong Huynh, Jinyu Yang, Ashish Tawari, Mubarak Shah, Son Tran, Raffay Hamid, Trishul Chilimbi, and Abhinav Shrivastava. 2025.
\newblock Collm: A large language model for composed image retrieval.
\newblock In \emph{Proceedings of the Computer Vision and Pattern Recognition Conference}, pages 3994--4004.

\bibitem[{Jian et~al.(2025)Jian, Zhang, Liang, Xie, He, Leng, and Yin}]{rzenembed}
Weijian Jian, Yajun Zhang, Dawei Liang, Chunyu Xie, Yixiao He, Dawei Leng, and Yuhui Yin. 2025.
\newblock Rzenembed: Towards comprehensive multimodal retrieval.
\newblock \emph{arXiv preprint arXiv:2510.27350}.

\bibitem[{Jiang et~al.(2024{\natexlab{a}})Jiang, Song, Zhang, Huang, Deng, Sun, Zhang, Wang, and Zhuang}]{e5v}
Ting Jiang, Minghui Song, Zihan Zhang, Haizhen Huang, Weiwei Deng, Feng Sun, Qi~Zhang, Deqing Wang, and Fuzhen Zhuang. 2024{\natexlab{a}}.
\newblock E5-v: Universal embeddings with multimodal large language models.
\newblock \emph{arXiv preprint arXiv:2407.12580}.

\bibitem[{Jiang et~al.(2024{\natexlab{b}})Jiang, Meng, Yang, Yavuz, Zhou, and Chen}]{vlm2vec}
Ziyan Jiang, Rui Meng, Xinyi Yang, Semih Yavuz, Yingbo Zhou, and Wenhu Chen. 2024{\natexlab{b}}.
\newblock Vlm2vec: Training vision-language models for massive multimodal embedding tasks.
\newblock \emph{arXiv preprint arXiv:2410.05160}.

\bibitem[{Karthik et~al.(2024)Karthik, Roth, Mancini, and Akata}]{cirevl}
Shyamgopal Karthik, Karsten Roth, Massimiliano Mancini, and Zeynep Akata. 2024.
\newblock Vision-by-language for training-free compositional image retrieval.
\newblock \emph{International Conference on Learning Representations (ICLR)}.

\bibitem[{Li et~al.(2023)Li, Li, Savarese, and Hoi}]{blip2}
Junnan Li, Dongxu Li, Silvio Savarese, and Steven Hoi. 2023.
\newblock Blip-2: Bootstrapping language-image pre-training with frozen image encoders and large language models.
\newblock In \emph{International conference on machine learning}, pages 19730--19742. PMLR.

\bibitem[{Li et~al.(2022)Li, Li, Xiong, and Hoi}]{blip}
Junnan Li, Dongxu Li, Caiming Xiong, and Steven Hoi. 2022.
\newblock Blip: Bootstrapping language-image pre-training for unified vision-language understanding and generation.
\newblock In \emph{International conference on machine learning}, pages 12888--12900. PMLR.

\bibitem[{Lin et~al.()Lin, Lee, Shoeybi, Lin, Catanzaro, and Ping}]{mmembed}
Sheng-Chieh Lin, Chankyu Lee, Mohammad Shoeybi, Jimmy Lin, Bryan Catanzaro, and Wei Ping.
\newblock Mm-embed: Universal multimodal retrieval with multimodal llms.
\newblock In \emph{The Thirteenth International Conference on Learning Representations}.

\bibitem[{Liu et~al.(2021)Liu, Rodriguez-Opazo, Teney, and Gould}]{cirr}
Zheyuan Liu, Cristian Rodriguez-Opazo, Damien Teney, and Stephen Gould. 2021.
\newblock Image retrieval on real-life images with pre-trained vision-and-language models.
\newblock In \emph{Proceedings of the IEEE/CVF International Conference on Computer Vision (ICCV)}, pages 2125--2134.

\bibitem[{{OpenSearch-AI}(2025)}]{ops}
{OpenSearch-AI}. 2025.
\newblock \href {https://huggingface.co/OpenSearch-AI/Ops-MM-embedding-v1-7B} {Opensearch-ai/ops-mm-embedding-v1-7b}.

\bibitem[{Psomas et~al.(2025)Psomas, Retsinas, Efthymiadis, Filntisis, Avrithis, Maragos, Chum, and Tolias}]{icir}
Bill Psomas, George Retsinas, Nikos Efthymiadis, Panagiotis Filntisis, Yannis Avrithis, Petros Maragos, Ondrej Chum, and Giorgos Tolias. 2025.
\newblock Instance-level composed image retrieval.
\newblock In \emph{The Thirty-ninth Annual Conference on Neural Information Processing Systems}.

\bibitem[{Saito et~al.(2023)Saito, Sohn, Zhang, Li, Lee, Saenko, and Pfister}]{pic2word}
Kuniaki Saito, Kihyuk Sohn, Xiang Zhang, Chun-Liang Li, Chen-Yu Lee, Kate Saenko, and Tomas Pfister. 2023.
\newblock Pic2word: Mapping pictures to words for zero-shot composed image retrieval.
\newblock In \emph{Proceedings of the IEEE/CVF Conference on Computer Vision and Pattern Recognition}, pages 19305--19314.

\bibitem[{Schuhmann et~al.(2021)Schuhmann, Vencu, Beaumont, Kaczmarczyk, Mullis, Katta, Coombes, Jitsev, and Komatsuzaki}]{laion400m}
Christoph Schuhmann, Richard Vencu, Romain Beaumont, Robert Kaczmarczyk, Clayton Mullis, Aarush Katta, Theo Coombes, Jenia Jitsev, and Aran Komatsuzaki. 2021.
\newblock Laion-400m: Open dataset of clip-filtered 400 million image-text pairs.
\newblock \emph{arXiv preprint arXiv:2111.02114}.

\bibitem[{Song et~al.(2025)Song, Lin, Wen, Hou, Xu, and Nie}]{song2025comprehensive}
Xuemeng Song, Haoqiang Lin, Haokun Wen, Bohan Hou, Mingzhu Xu, and Liqiang Nie. 2025.
\newblock A comprehensive survey on composed image retrieval.
\newblock \emph{ACM Transactions on Information Systems}, 44(1):1--54.

\bibitem[{Vaze et~al.(2023)Vaze, Carion, and Misra}]{genecis}
Sagar Vaze, Nicolas Carion, and Ishan Misra. 2023.
\newblock Genecis: A benchmark for general conditional image similarity.
\newblock In \emph{Proceedings of the IEEE/CVF Conference on Computer Vision and Pattern Recognition}, pages 6862--6872.

\bibitem[{Ventura et~al.(2024)Ventura, Yang, Schmid, and Varol}]{covr2}
Lucas Ventura, Antoine Yang, Cordelia Schmid, and G{\"u}l Varol. 2024.
\newblock {CoVR-2}: Automatic data construction for composed video retrieval.
\newblock \emph{IEEE TPAMI}.

\bibitem[{Wan et~al.(2025)Wan, Zou, and Zhang}]{wan2025composed}
Yongquan Wan, Guobing Zou, and Bofeng Zhang. 2025.
\newblock Composed image retrieval: a survey on recent research and development.
\newblock \emph{Applied Intelligence}, 55(6):482.

\bibitem[{Wang et~al.(2024)Wang, Bai, Tan, Wang, Fan, Bai, Chen, Liu, Wang, Ge et~al.}]{qwen2vl}
Peng Wang, Shuai Bai, Sinan Tan, Shijie Wang, Zhihao Fan, Jinze Bai, Keqin Chen, Xuejing Liu, Jialin Wang, Wenbin Ge, et~al. 2024.
\newblock Qwen2-vl: Enhancing vision-language model's perception of the world at any resolution.
\newblock \emph{arXiv preprint arXiv:2409.12191}.

\bibitem[{Wu et~al.(2025)Wu, Li, Zhou, Lin, Gao, Yan, ming Yin, Bai, Xu, Chen, Chen, Tang, Zhang, Wang, Yang, Yu, Cheng, Liu, Li, Zhang, Meng, Wei, Ni, Chen, Cao, Peng, Qu, Wu, Wang, Yu, Wen, Feng, Xu, Wang, Zhang, Zhu, Wu, Cai, and Liu}]{qwenedit}
Chenfei Wu, Jiahao Li, Jingren Zhou, Junyang Lin, Kaiyuan Gao, Kun Yan, Sheng ming Yin, Shuai Bai, Xiao Xu, Yilei Chen, Yuxiang Chen, Zecheng Tang, Zekai Zhang, Zhengyi Wang, An~Yang, Bowen Yu, Chen Cheng, Dayiheng Liu, Deqing Li, Hang Zhang, Hao Meng, Hu~Wei, Jingyuan Ni, Kai Chen, Kuan Cao, Liang Peng, Lin Qu, Minggang Wu, Peng Wang, Shuting Yu, Tingkun Wen, Wensen Feng, Xiaoxiao Xu, Yi~Wang, Yichang Zhang, Yongqiang Zhu, Yujia Wu, Yuxuan Cai, and Zenan Liu. 2025.
\newblock \href {http://arxiv.org/abs/2508.02324} {Qwen-image technical report}.

\bibitem[{Wu et~al.(2021)Wu, Gao, Guo, Al-Halah, Rennie, Grauman, and Feris}]{fashioniq}
Hui Wu, Yupeng Gao, Xiaoxiao Guo, Ziad Al-Halah, Steven Rennie, Kristen Grauman, and Rogerio Feris. 2021.
\newblock Fashion iq: A new dataset towards retrieving images by natural language feedback.
\newblock In \emph{Proceedings of the IEEE/CVF Conference on computer vision and pattern recognition}, pages 11307--11317.

\bibitem[{Wu et~al.(2024)Wu, Lin, and Yang}]{weimocir}
Ren-Di Wu, Yu-Yen Lin, and Huei-Fang Yang. 2024.
\newblock Training-free zero-shot composed image retrieval via weighted modality fusion and similarity.
\newblock In \emph{International Conference on Technologies and Applications of Artificial Intelligence}, pages 77--90. Springer.

\bibitem[{Yang et~al.(2025)Yang, Li, Yang, Zhang, Hui, Zheng, Yu, Gao, Huang, Lv et~al.}]{qwen3}
An~Yang, Anfeng Li, Baosong Yang, Beichen Zhang, Binyuan Hui, Bo~Zheng, Bowen Yu, Chang Gao, Chengen Huang, Chenxu Lv, et~al. 2025.
\newblock Qwen3 technical report.
\newblock \emph{arXiv preprint arXiv:2505.09388}.

\bibitem[{Yang et~al.(2020)Yang, Song, Han, Wen, Nie, and Nie}]{yang2020generative}
Xin Yang, Xuemeng Song, Xianjing Han, Haokun Wen, Jie Nie, and Liqiang Nie. 2020.
\newblock Generative attribute manipulation scheme for flexible fashion search.
\newblock In \emph{Proceedings of the 43rd international acm sigir conference on research and development in information retrieval}, pages 941--950.

\bibitem[{Yang et~al.(2024)Yang, Xue, Qian, Dong, and Xu}]{ldre}
Zhenyu Yang, Dizhan Xue, Shengsheng Qian, Weiming Dong, and Changsheng Xu. 2024.
\newblock Ldre: Llm-based divergent reasoning and ensemble for zero-shot composed image retrieval.
\newblock In \emph{Proceedings of the 47th International ACM SIGIR conference on research and development in information retrieval}, pages 80--90.

\bibitem[{Zhang et~al.()Zhang, Luan, Hu, Lee, Qiao, Chen, Su, and Chang}]{magiclens}
Kai Zhang, Yi~Luan, Hexiang Hu, Kenton Lee, Siyuan Qiao, Wenhu Chen, Yu~Su, and Ming-Wei Chang.
\newblock Magiclens: Self-supervised image retrieval with open-ended instructions.
\newblock In \emph{Forty-first International Conference on Machine Learning}.

\bibitem[{Zhang et~al.(2024)Zhang, Zhang, Xie, Li, Dai, Long, Xie, Zhang, Li, and Zhang}]{gme}
Xin Zhang, Yanzhao Zhang, Wen Xie, Mingxin Li, Ziqi Dai, Dingkun Long, Pengjun Xie, Meishan Zhang, Wenjie Li, and Min Zhang. 2024.
\newblock Gme: Improving universal multimodal retrieval by multimodal llms.
\newblock \emph{arXiv preprint arXiv:2412.16855}.

\bibitem[{Zhou et~al.(2024)Zhou, Liu, Xiao, Zhao, and Xiong}]{vista}
Junjie Zhou, Zheng Liu, Shitao Xiao, Bo~Zhao, and Yongping Xiong. 2024.
\newblock Vista: Visualized text embedding for universal multi-modal retrieval.
\newblock \emph{arXiv preprint arXiv:2406.04292}.

\bibitem[{Zhou et~al.(2025)Zhou, Xiong, Liu, Liu, Xiao, Wang, Zhao, Zhang, and Lian}]{megapairs}
Junjie Zhou, Yongping Xiong, Zheng Liu, Ze~Liu, Shitao Xiao, Yueze Wang, Bo~Zhao, Chen~Jason Zhang, and Defu Lian. 2025.
\newblock Megapairs: Massive data synthesis for universal multimodal retrieval.
\newblock In \emph{Proceedings of the 63rd Annual Meeting of the Association for Computational Linguistics (Volume 1: Long Papers)}, pages 19076--19095.

\bibitem[{Zhu et~al.(2025)Zhu, Wang, Chen, Liu, Ye, Gu, Tian, Duan, Su, Shao et~al.}]{internvl3}
Jinguo Zhu, Weiyun Wang, Zhe Chen, Zhaoyang Liu, Shenglong Ye, Lixin Gu, Hao Tian, Yuchen Duan, Weijie Su, Jie Shao, et~al. 2025.
\newblock Internvl3: Exploring advanced training and test-time recipes for open-source multimodal models.
\newblock \emph{arXiv preprint arXiv:2504.10479}.

\end{thebibliography}

\appendix

\clearpage

\begin{figure*}[b]
    \centering
    \includegraphics[width=0.9\textwidth]{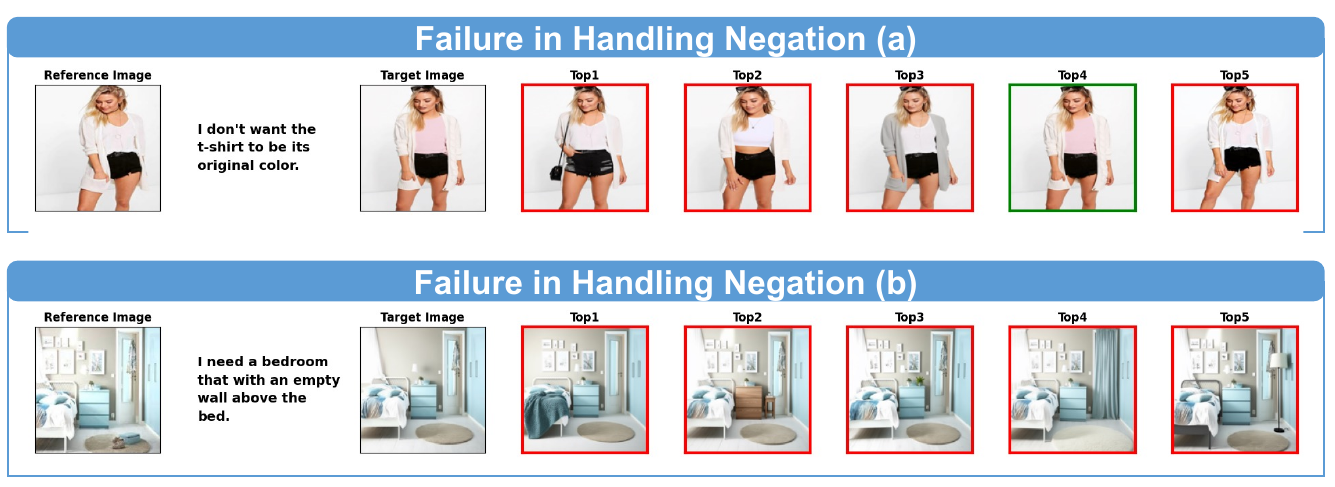}
    \caption{Example of Error Type: \emph{Failure in Handling Negation}}
    \label{fig-appendix-error-analysis-negation}
\end{figure*} 
\section{\ours}

\subsection{Details of Construction}
\label{appendix-edir-construction}
As shown in~\autoref{fig-data-synth-pipeline}, we use Qwen25-VL-32B-Instruct~\cite{qwen25vl} for both seed image selection and edit-instruction generation.
For image editing, we use Qwen-Image-Edit~\cite{qwenedit} (version Qwen-Image-Edit-2509) to generate the target images.
For query rewriting, we use Qwen3-32B~\cite{qwen3} to rewrite each edit instruction into a \cir query according to a predefined template.
We adopt two rewriting strategies: (i) directly rewriting the instruction into a \cir query, and (ii) rewriting it into a negation-form query.
Since not all categories are suitable for negation, we only apply negation rewriting to the \textit{color}, \textit{shape}, \textit{material}, \textit{texture}, \textit{style}, \textit{weather}, and \textit{time} categories. 
 
\subsection{Error Analysis}
\label{appendix-error-analysis}
For error analysis, we examine representative examples where the state-of-the-art model, RzenEmbed-7B, achieved a low Recall@1 score. 
\paragraph{Failure in Handling Negation. }
As shown in~\autoref{fig-appendix-error-analysis-negation}, we observe two types of negation-related queries.
The first type is explicit negation, where the user requests \emph{not} to keep an attribute of the reference image (\eg not keeping the T-shirt in its original color), as shown in~\autoref{fig-appendix-error-analysis-negation}(a).
The second type corresponds to \textit{remove} edits, where the user requests an object or region to be removed (\eg an empty wall above the bed), as shown in~\autoref{fig-appendix-error-analysis-negation}(b).
In both cases, the retrieved results tend to preserve the negated attribute or fail to realize the removal, indicating difficulty in mapping negation to the intended target state.

\begin{figure*}[!h]
    \centering
    \includegraphics[width=0.9\textwidth]{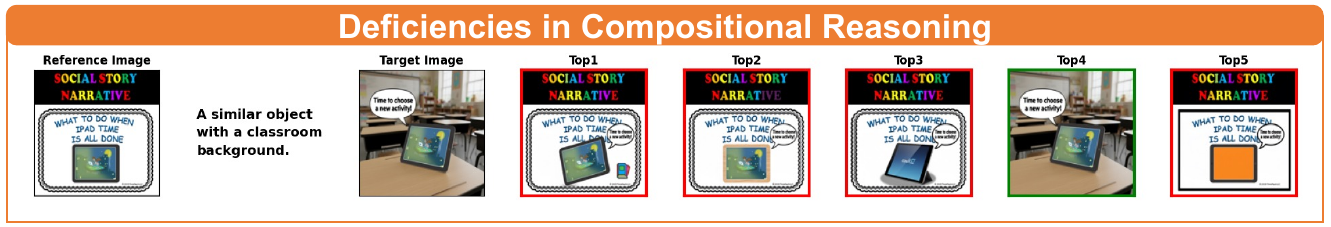}
    \caption{Example of Error Type: \emph{Deficiencies in Compositional Reasoning}}
    \label{fig-appendix-error-analysis-reasoning}
\end{figure*}

\begin{figure*}[!h]
    \centering
    \includegraphics[width=0.9\textwidth]{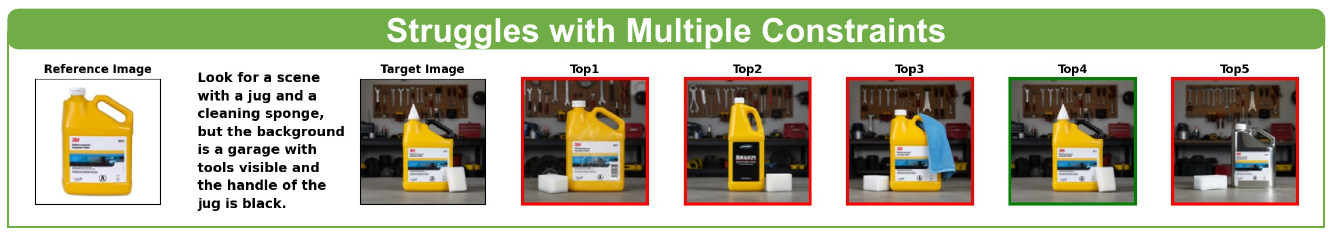}
    \caption{Example of Error Type: \emph{Struggles with Multiple Constraints}}
    \label{fig-appendix-error-analysis-constraints}
\end{figure*}

\paragraph{Deficiencies in Compositional Reasoning. }
Models exhibit poor performance on categories such as \textit{count}, \textit{spatial}, \textit{style}, and \textit{viewpoint}, which require compositional reasoning.
As shown in~\autoref{fig-appendix-error-analysis-reasoning}, the query asks for a similar object \emph{with a classroom background}.
However, the retrieved images often match the object appearance while failing to align the global scene context, suggesting limited capability in jointly reasoning about foreground content and background context.
\paragraph{Struggles with Multiple Constraints. }
In the \textit{complex} category, queries specify multiple constraints, yet models frequently retrieve images that only partially satisfy them.
As shown in~\autoref{fig-appendix-error-analysis-constraints}, the top retrieved result matches the presence of ``a jug'' and ``a sponge'' and roughly matches ``the garage-like background'',
but fails to satisfy the fine attribute constraint that ``the jug handle is black''.
This indicates a weakness in composing and verifying multiple distinct requirements from a single query.

\begin{figure*}[!h]
    \centering
    \includegraphics[width=0.9\textwidth]{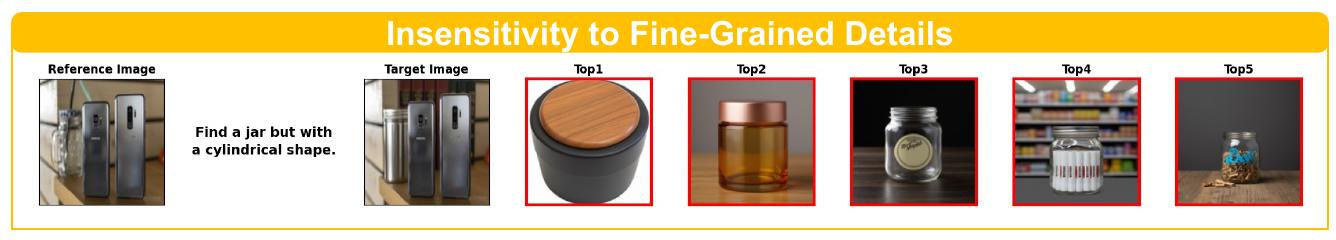}
    \caption{Example of Error Type: \emph{Insensitivity to Fine-Grained Details}}
    \label{fig-appendix-error-analysis-finegrained}
\end{figure*}

\paragraph{Insensitivity to Fine-Grained Details. }
For categories such as \textit{texture}, \textit{material}, and \textit{shape}, the distinctions between the source and target images can be subtle, and current models tend to overlook such fine-grained visual cues.
As shown in~\autoref{fig-appendix-error-analysis-finegrained}, the model ignores the fine-grained details of the jar in the reference image and retrieves results that merely contain a jar, without preserving the intended subtle characteristics.


\FloatBarrier 
\section{Experiment Settings}
\subsection{Evaluation Details} 
\label{appendix-evaluation-details}

\paragraph{Model Settings} We provide the details of the evaluated models in ~\autoref{appendix-model-configuration}. 
For Non-MLLM-based models, we use their CLIP-L/14 variants to ensure a fair comparison, including \picword, \textsc{SEARLE} and \magiclens. 
For MLLM-based models, we set the maximum sequence length to 2048 and the maximum number of pixels to 1280. And the instruction we used is ``Given an image, find a similar image satisfying the query. ''. 
\begin{table*}[htbp]
\centering
\footnotesize
\resizebox{0.90\textwidth}{!}{%
\begin{tabular}{lllcrr}
\toprule
\textbf{Model} & \textbf{Release} & \textbf{Version} & \textbf{Pretrained} & \textbf{Backbone} \\
\midrule
\multicolumn{6}{c}{\emph{\textbf{MLLM-based Models}}} \\
 \midrule

RzenEmbed-7B & 2025-11 & \texttt{RzenEmbed-7B-V2} & \xmark & Qwen2-VL-7B \\ 
Ops-embedding & 2025-07 & \texttt{Ops-MM-Embedding-v1-7B} & - & Qwen2-VL-7B \\ 
GME-2B & 2024-12 & \texttt{gme-Qwen2-VL-2B-Instruct} & \checkmark & Qwen2-VL-2B \\
GME-7B & 2024-12 & \texttt{gme-Qwen2-VL-7B-Instruct} & \checkmark & Qwen2-VL-7B \\
MMRet-MLLM & 2025-04 &  \texttt{BGE-VL-MLLM-S1} & \checkmark & Llava-Mistral-7B \\ 
E5-V & 2024-07 & \texttt{e5-v}& \xmark & Llava-llama3-8B \\
VLM2Vec-2B & 2025-05 & \texttt{VLM2Vec-V2.0} & \checkmark & Qwen2-VL-2B  \\
UniME-2B & 2025-10 & \texttt{UniME-V2-Qwen2VL-2B} & \checkmark & Qwen2-VL-2B \\
UniME-7B & 2025-10 & \texttt{UniME-V2-Qwen2VL-7B} & \checkmark & Qwen2-VL-7B \\
mmE5 & 2025-02 & \texttt{mmE5-mllama-11b-instruct}  & \xmark & Llama-3.2-Vision\\ 
\midrule
\multicolumn{6}{c}{\emph{\textbf{Non-MLLM-based Models}}} \\
 \midrule
\picword  & 2023-02 & \texttt{PIC2WORD(CLIP-L/14)} & \xmark & CLIP-L/14 \\
\searle & 2023-03 & \texttt{SEARLE(CLIP-L/14)} & \xmark & CLIP-L/14 \\
\magiclens & 2024-03 & \texttt{MAGICLENS(CLIP-L/14)} & \xmark & CLIP-L/14 \\
\bottomrule
\end{tabular}
}
\caption{ Details of the evaluated multimodal embedding models in \ours. }
\label{appendix-model-configuration}
\end{table*}

\paragraph{Benchmark Settings} 
We evaluate models on \ours using Recall@1. 
For the other benchmarks, we follow their standard evaluation metrics: \cirr (Recall@1), \circo (mAP@5), \fashioniq (Recall@10), and \genecis (Recall@1). 
For \cirr, we follow the evaluation protocol in~\cite{e5v}, excluding the reference image from the retrieval corpus. For both \cirr and \circo, we report the results on the validation set.

\subsection{Training Settings } 
\label{appendix-training-settings}

Using our data synthesis pipeline, we edit 500,000 images from LAION-400M, producing 1,087,710 training instances. Each instance consists of a reference image, a query, a target image, and three hard negatives sampled from the same source. From this pool, we sample 15,000 triplets per category across 15 categories, yielding a final training set of 225,000 instances.
We train Qwen2.5-VL-7B-Instruct~\cite{qwen25vl} with a batch size of 128. The maximum number of image tokens is set to 1,280, and the maximum sequence length is 1,500. The learning rate is 3e-5 with a weight decay of 0.01. We apply LoRA only to the q\_proj, k\_proj, v\_proj, up\_proj, down\_proj, and gate\_proj layers. Training uses an InfoNCE-style loss with a temperature of 0.03.

\subsection{Results} 
We provide further details on model performance on \ours using additional metrics (i.e., Recall@3), as shown in \autoref{table-appendix-main-results}. All models exhibit a significant performance increase when evaluated with Recall@3. However, the average performance of MLLM-based models remains close to 60, indicating a substantial gap that still needs to be addressed. In addition, the zero-shot models (e.g., MMRet-MLLM, E5-V, and \magiclens) still fail to perform well.
Moreover, our benchmark aims to evaluate the fine-grained capabilities of these models. As shown in \autoref{table-appendix-main-results}, \ours can still reveal model weaknesses in specific categories, such as \textit{remove} and \textit{view}.

\begin{table*}[htbp]
\centering
\small
\footnotesize
\renewcommand{\arraystretch}{1.1}
\setlength{\tabcolsep}{2pt}
\scalebox{0.9}{
\begin{tabular}{lcccccccccccccccc}
\toprule[.1em]
\multirow{2}{*}{\textbf{Metric}} & \multirow{2}{*}{\textbf{Total}} & \multicolumn{4}{c}{\textbf{Attribute}} & \multicolumn{4}{c}{\textbf{Object}} & \multicolumn{3}{c}{\textbf{Relationship}} &  \multicolumn{3}{c}{\textbf{Style}} & \textbf{Complex} \\
 \cmidrule(lr){3-6}  \cmidrule(lr){7-10} \cmidrule(lr){11-13} \cmidrule(lr){14-16} \cmidrule(lr){17-17} 
 & & Color & Material & Shape & Texture & Add & Remove & Replace & Count & Spatial & Action & View & Style & Weather & Time & Complex \\
\midrule
\multicolumn{16}{c}{\emph{\textbf{Non MLLM-based Models}}} \\
\midrule
\picword & \textbf{42.2} & \textbf{42.7} & \textbf{34.3} & \textbf{37.3} & \textbf{38.0} & \textbf{50.0} & \textbf{31.0} & 40.0 & \textbf{47.7} & \textbf{35.7} & \textbf{50.0} & \textbf{32.0} & \textbf{42.3} & \textbf{49.3} & \textbf{51.0} & \textbf{45.9} \\
\searle & 33.0 & 34.0 & 29.0 & 28.0 & 26.3 & 39.0 & 19.7 & 36.0 & 40.3 & 24.7 & 48.0 & 22.0 & 21.7 & 42.3 & 37.0 & 38.5 \\
\magiclens & 29.9 & 29.7 & 17.7 & 17.7 & 14.7 & 37.0 & 28.7 & \textbf{41.3} & 31.7 & 33.0 & 38.3 & 23.7 & \textbf{42.3} & 21.0 & 34.3 & 32.5 \\
\noalign{\vskip 0.5ex}\hdashline\noalign{\vskip 0.5ex}
Avg. & 35.0 & 35.4 & 27.0 & 27.7 & 26.3 & 42.0 & 26.4 & 39.1 & 39.9 & 31.1 & 45.4 & 25.9 & 35.4 & 37.6 & 40.8 & 39.0 \\
\midrule
\multicolumn{16}{c}{\emph{\textbf{MLLM-based Models}}} \\
\midrule
Ops-embedding & \textbf{71.3} & \textbf{70.3} & \textbf{65.7} & \textbf{64.3} & \textbf{70.0} & 86.7 & 49.7 & 81.7 & \textbf{73.7} & \textbf{69.3} & \textbf{81.7} & 56.0 & \textbf{70.0} & \textbf{70.0} & \textbf{76.3} & \textbf{76.4} \\
RzenEmbed-7B & 69.6 & 64.3 & 57.7 & 63.3 & 60.7 & \textbf{87.0} & \textbf{57.7} & \textbf{84.0} & 71.0 & \textbf{69.3} & 80.3 & \textbf{57.0} & 67.7 & 66.3 & 74.0 & 74.8 \\
GME-2B & 66.1 & 61.0 & 59.7 & 61.7 & 62.7 & 79.7 & 48.7 & 75.3 & 68.3 & 64.3 & 75.7 & 48.3 & 66.7 & 68.7 & 71.7 & 71.0 \\
GME-7B & 62.9 & 59.0 & 57.0 & 58.3 & 53.0 & 77.7 & 52.3 & 75.3 & 65.0 & 60.3 & 68.3 & 50.3 & 58.7 & 63.7 & 68.7 & 67.9 \\
VLM2Vec-2B & 61.7 & 59.7 & 52.3 & 61.7 & 67.0 & 76.3 & 49.3 & 60.7 & 68.7 & 61.0 & 68.7 & 45.7 & 51.3 & 58.3 & 66.0 & 68.0 \\
MMRet-MLLM & 58.0 & 53.3 & 44.7 & 49.3 & 49.3 & 75.0 & 48.3 & 68.3 & 57.7 & 63.7 & 62.7 & 46.0 & 41.7 & 57.7 & 63.0 & 69.8 \\
E5-V & 56.2 & 47.0 & 47.0 & 48.0 & 54.0 & 70.7 & 37.3 & 69.3 & 61.7 & 60.7 & 70.0 & 36.7 & 56.7 & 50.0 & 59.7 & 63.1 \\
mmE5 & 53.8 & 53.3 & 44.3 & 48.0 & 49.3 & 64.3 & 48.3 & 64.0 & 56.0 & 56.0 & 62.3 & 37.7 & 53.7 & 52.0 & 54.7 & 57.1 \\
UniME-7B & 50.2 & 36.3 & 29.3 & 37.7 & 40.0 & 64.3 & 38.7 & 56.3 & 60.7 & 56.0 & 66.0 & 37.0 & 48.7 & 46.3 & 51.0 & 63.4 \\
UniME-2B & 49.4 & 45.7 & 39.7 & 40.7 & 46.3 & 59.7 & 40.7 & 67.0 & 51.0 & 48.3 & 61.7 & 38.7 & 41.3 & 41.0 & 51.3 & 56.1 \\
\noalign{\vskip 0.5ex}\hdashline\noalign{\vskip 0.5ex}
Avg. & 59.9 & 55.0 & 49.7 & 53.3 & 55.2 & 74.1 & 47.1 & 70.2 & 63.4 & 60.9 & 69.7 & 45.3 & 55.6 & 57.4 & 63.6 & 66.8 \\ 
\ourmodel & 80.8 & 76.3 & 79.0 & 73.3 & 81.3 & 92.7 & 66.7 & 87.3 & 82.0 & 73.0 & 87.7 & 63.7 & 86.0 & 89.7 & 87.7 & 82.4 \\
\bottomrule[.1em]
\end{tabular}
}
\caption{Models Recall@3 performances on \ours. } 
\label{table-appendix-main-results}
\end{table*}

\FloatBarrier 
\section{Prompts. }
\label{appendix-prompts}

As shown in ~\autoref{fig-data-synth-pipeline}, we first prompt Qwen25-VL-32B-Instruct to filter out the images that are not suitable for editing. The prompt is shown in ~\autoref{fig-judge-seed-image}. 
After obtaining the seed images, we prompt Qwen25-VL-32B-Instruct to generate edit instructions for these seed images. 
For each image, the MLLM is required to generate edit instructions for 5-6 categories and 3 edit instructions for each category. The prompt is shown in ~\autoref{fig-edit-inst-gen}. 
As we have two methods for prompt rewriting, we provide the prompt for direct rewriting in ~\autoref{fig-cir-query-generation}, and we provide the query negation rewrite prompt as shown in ~\autoref{fig-cir-negation-query-generation}. 
For the two stage filtering, we utilize the same prompt template, as shown in ~\autoref{fig-image-pair-matching}.


\begin{figure*}[htbp]
\centering
\begin{tcolorbox}[
    colback=black!5,
    colframe=black!75,
    fonttitle=\bfseries,
    title=Seed Image Selection,
    width=\textwidth,
    arc=2mm,
    boxrule=0.5pt,
    enhanced,
    left=6pt,           
    right=6pt,          
    top=4pt,            
    bottom=4pt,         
    attach boxed title to top center={yshift=-2mm},
    boxed title style={
        size=normal,
        colback=white,
        colframe=black!75, 
        coltext=black,
        arc=1mm,
    },
    coltitle=black,
    titlerule=0.5pt,
    title style={top color=white, bottom color=white}
]
\footnotesize
\setlength{\baselineskip}{1.1\baselineskip}

You are an AI assistant that judges if an image is suitable for common image editing tasks like adding/removing objects, replacing elements, or changing the background.

Analyze the provided image and determine its suitability.
\vspace{0.3em}

An image is considered \textbf{NOT suitable} if it is:
\begin{enumerate}[topsep=0pt, itemsep=2pt, partopsep=0pt, parsep=0pt]
    \item \textbf{Primarily Text:} A screenshot of a document, a presentation slide, or code with no significant visual elements.
    \item \textbf{Too Simple:} A solid color, a simple gradient, or a basic pattern with no distinct objects to manipulate.
    \item \textbf{Poor Quality:} The image is low-resolution, blurry, or heavily pixelated, especially when the composition is complex. This combination makes it impossible to identify or edit objects cleanly.
    \item \textbf{Too Abstract or Cluttered:} An abstract pattern, a dense texture, or a chaotic collage where there is no clear subject or distinction between foreground and background.
    \item \textbf{Functional:} A QR code, barcode, or captcha, where editing would destroy its purpose.
\end{enumerate}
\vspace{0.3em} 

Based on your analysis, provide your output ONLY in the following JSON format:
{\ttfamily
\noindent
\begin{tabbing}
\qquad\=\qquad\=\kill
\{\\
\>\textquotedbl useful\textquotedbl:\ <true\_or\_false>,\\
\>\textquotedbl reason\textquotedbl:\ \textquotedbl<A brief explanation for your decision.>\textquotedbl\\
\}\\
\end{tabbing}
}
\end{tcolorbox}
\caption{Prompt used to judge whether an image is suitable for image editing.}
\label{fig-judge-seed-image} 
\end{figure*}


\begin{figure*}[htbp]
\centering
\begin{tcolorbox}[
    colback=black!5,
    colframe=black!75,
    fonttitle=\bfseries,
    title=Edit Instruction Generation,
    width=\textwidth,
    arc=2mm,
    boxrule=0.5pt,
    enhanced,
    left=6pt,           
    right=6pt,          
    top=4pt,            
    bottom=4pt,         
    attach boxed title to top center={yshift=-2mm},
    boxed title style={
        size=normal,
        colback=white,
        colframe=black!75, 
        coltext=black,
        arc=1mm,
    },
    coltitle=black,
    titlerule=0.5pt,
    title style={top color=white, bottom color=white}
]
\footnotesize
\setlength{\baselineskip}{1.1\baselineskip}

The model must output a single valid JSON object with the following structure:
\vspace{0.3em}

{\ttfamily
\begin{tabbing}
\qquad\=\qquad\=\qquad\=\qquad\=\kill
\{ \\
\> "image\_description": "<one-sentence description of the image>", \\
\> "categories": \{ \\
\>\> "<category\_1>": \{ \\
\>\>\> "instructions": [ \\
\>\>\>\> "<instruction\_1>", \\
\>\>\>\> "<instruction\_2>", \\
\>\>\>\> "<instruction\_3>" \\
\>\>\> ] \\
\>\> \}, \\
\>\> "<category\_2>": \{ "instructions": [ ... ] \}, \\
\>\> ... \\
\> \} \\
\}
\end{tabbing}
}
\vspace{0.3em}

\textbf{--- RULES ---}
\begin{enumerate}[topsep=0pt, itemsep=2pt, partopsep=0pt, parsep=0pt]
    \item \textbf{Top-level keys:} The JSON root \emph{must} contain exactly two keys: \texttt{"image\_description"} and \texttt{"categories"}.
    \item \textbf{Categories count:} The \texttt{"categories"} object must contain 5--6 keys, each selected from the allowed category list.
    \item \textbf{Allowed category keys:} \\
          \texttt{color}, \texttt{material}, \texttt{shape}, \texttt{texture}, \texttt{addition}, \texttt{remove}, \texttt{replace}, \texttt{cardinality}, \texttt{spatial}, \texttt{action}, \texttt{viewpoint}, \texttt{style}, \texttt{time}, \texttt{weather}.
    \item \textbf{Instructions list:} Each chosen category must contain an \texttt{"instructions"} list with 2--3 atomic editing instructions.
    \item \textbf{Instruction independence:} Instructions across different categories must be combinable without logical conflicts. Do not create edits that negate each other (e.g., removing an object and also recoloring it).
    \item \textbf{Atomicity:} Each instruction must describe a single concrete change applied to one object or one cohesive group.
    \item \textbf{Real-world plausibility:} All edits must be realistic and physically plausible; avoid fantasy-like transformations.
    \item \textbf{Concreteness:} Avoid vague terms like ``enhance'' or ``improve''; instead, specify explicit changes (e.g., ``Change the sky to a clear, bright blue.'').
    \item \textbf{Category balance:} Pay particular attention to \texttt{remove}, \texttt{replace}, \texttt{cardinality}, \texttt{viewpoint}, \texttt{shape}, \texttt{time}, and \texttt{texture}, ensuring these are used and not neglected.
\end{enumerate}
\vspace{0.3em}
\textbf{--- EXAMPLE ---}
\vspace{0.3em}

{\ttfamily
\begin{tabbing}
\qquad\=\qquad\=\qquad\=\qquad\=\kill 
\{ \\
\> "image\_description": \begin{minipage}[t]{0.7\textwidth}\strut"A woman wearing a dress standing in a living room."\strut\end{minipage}, \\
\> "categories": \{ \\
\>\> "remove": \{ \\
\>\>\> "instructions": [ \\
\>\>\>\> \begin{minipage}[t]{0.55\textwidth}\strut"Remove the coffee table from the scene.",\strut\end{minipage} \\
\>\>\>\> \begin{minipage}[t]{0.55\textwidth}\strut"Remove the rug from under the furniture, exposing the floor.",\strut\end{minipage} \\
\>\>\>\> \begin{minipage}[t]{0.55\textwidth}\strut"Remove the woman, leaving an empty living room."\strut\end{minipage} \\
\>\>\> ] \\
\>\> \}, \\
\> \} \\
\}
\end{tabbing}
}

\end{tcolorbox}
\caption{Prompt used to generate image editing instructions.}
\label{fig-edit-inst-gen}
\end{figure*}

\begin{figure*}[htbp]
\centering
\begin{tcolorbox}[
    colback=black!5,
    colframe=black!75,
    fonttitle=\bfseries,
    title=CIR Query Generation,
    width=\textwidth,
    arc=2mm,
    boxrule=0.5pt,
    enhanced,
    left=6pt,           
    right=6pt,          
    top=4pt,            
    bottom=4pt,         
    attach boxed title to top center={yshift=-2mm},
    boxed title style={
        size=normal,
        colback=white,
        colframe=black!75, 
        coltext=black,
        arc=1mm,
    },
    coltitle=black,
    titlerule=0.5pt,
    title style={top color=white, bottom color=white}
]
\footnotesize
\setlength{\baselineskip}{1.1\baselineskip}

Given an image edit query, rewrite it into an image search query. The goal is to create a search that finds an image matching the final, desired scene.

\textbf{--- GUIDELINES ---}
\begin{enumerate}
    \item \textbf{Describe the Final State:} Convert action commands (like "add", "make", "move") into descriptive phrases ("a picture of...", "a scene where...").
    \item \textbf{Omit Comparative Words:} Always remove words that compare to the original image, like "larger", "more", "brighter".
    \item \textbf{Handle Relational Details Intelligently (CRITICAL):}
        \begin{enumerate}
            \item If adding a NEW object: You can often omit its location relative to existing objects to get a better search. Focus on the new object itself. (e.g., "Add a bird on the fence" -> "A picture with a bird").
            \item If changing the relationship between EXISTING objects: The new relationship is the most important detail and MUST be included in the search. (e.g., "Move the cat onto the sofa" -> "A picture of a cat on a sofa").
        \end{enumerate}
\end{enumerate}

\textbf{--- EXAMPLES ---}

\textbf{CASE 1: Adding a new object (Omit relation)} \\
\textit{Edit Query:} Add a flock of seagulls flying near the kitesurfer. \\
\textit{Rewritten Search:} I want to see a picture with seagulls flying. \\
\textit{(Reason: The core request is to add seagulls. Their exact position 'near the kitesurfer' is secondary and omitted.)}
\vspace{1em}

\textbf{CASE 2: Changing an object’s relationship (Keep relation)} \\
\textit{Edit Query:} Move the dog so it is sitting at the man's feet. \\
\textit{Rewritten Search:} A picture of a dog sitting at a man's feet. \\
\textit{(Reason: The entire point of the edit is the new relationship between the dog and the man. This detail is essential and must be kept.)}
\vspace{1em}

\textbf{CASE 3: Changing a scene attribute (Omit comparative)} \\
\textit{Edit Query:} Make it a windy day with larger waves. \\
\textit{Rewritten Search:} I want to see a windy weather of this place. \\
\textit{(Reason: The comparative "larger" is omitted. The core idea is the windy weather.)}

\vspace{1em}
\textbf{--- TASK ---} \\
Query: [FILL\_THE\_QUERY] \\
You only need to output the rewritten query without any other words or characters. The output should begin with the following prefix. Here is the prefix: [FILL\_THE\_PREFIX] \\
If the prefix is "empty", you should simply return a description of the final scene.

\end{tcolorbox}
\caption{Prompt used to convert edit instruction to CIR query. This prompt corresponds to the direct rewriting strategy.}
\label{fig-cir-query-generation}
\end{figure*}

\begin{figure*}[htbp]
\centering
\begin{tcolorbox}[
    colback=black!5,
    colframe=black!75,
    fonttitle=\bfseries,
    title=CIR Negation Query,
    width=\textwidth,
    arc=2mm,
    boxrule=0.5pt,
    enhanced,
    attach boxed title to top center={yshift=-2mm},
    boxed title style={
        size=normal,
        colback=white,
        colframe=black!75, 
        coltext=black,
        arc=1mm,
    },
    coltitle=black,
    titlerule=0.5pt,
    title style={top color=white, bottom color=white}
]
\footnotesize
\setlength{\baselineskip}{1.1\baselineskip}


Given an image edit query, rewrite it into an image search query. The goal is to create a search that finds an image matching the final, desired scene.

\textbf{--- GUIDELINES ---}
\begin{enumerate}[topsep=0pt, itemsep=2pt, partopsep=0pt, parsep=0pt]
    \item Convert positive statements about an attribute into a negative or relative query.
    \item Focus on the attribute being changed, not the final state.
    \item Avoid describing the final appearance; instead, state what should be different.
\end{enumerate}

\textbf{--- EXAMPLES ---}

\textbf{CASE 1: Changing an attribute (Color)} \\
\textit{Edit Query:} Change the dress color to red. \\
\textit{Rewritten Search:} Find this dress but in a different color. \\
\textit{(Reason: The query asks for any color other than the original, not specifically red.)}
\vspace{1em}

\textbf{CASE 2: Changing an attribute (Style)} \\
\textit{Edit Query:} Change the style to a watercolor painting. \\
\textit{Rewritten Search:} Show me this picture but not as a photograph. \\
\textit{(Reason: The query negates the current style to find alternatives.)}
\vspace{1em}

\textbf{--- TASK ---} \\
Query: [FILL\_THE\_QUERY] \\
You only need to output the rewritten query without any other words or characters. The output should begin with the following prefix. Here is the prefix: [FILL\_THE\_PREFIX] \\
If the prefix is "empty", you should simply return a description of the final scene.

\end{tcolorbox}
\caption{Prompt used to convert an edit instruction to a negation \cir query.}
\label{fig-cir-negation-query-generation}
\end{figure*}
 

\begin{figure*}[htbp]
\centering
\begin{tcolorbox}[
    colback=black!5,
    colframe=black!75,
    fonttitle=\bfseries,
    title=Image Pair Matching,
    width=\textwidth,
    arc=2mm,
    boxrule=0.5pt,
    enhanced,
    left=6pt,           
    right=6pt,          
    top=4pt,            
    bottom=4pt,         
    attach boxed title to top center={yshift=-2mm},
    boxed title style={
        size=normal,
        colback=white,
        colframe=black!75, 
        coltext=black,
        arc=1mm,
    },
    coltitle=black,
    titlerule=0.5pt,
    title style={top color=white, bottom color=white}
]
\footnotesize
\setlength{\baselineskip}{1.1\baselineskip}

\textbf{Task:} \\
Your task is to act as a quality control checkpoint. You will be given a \texttt{source image}, a \texttt{text description}, and a \texttt{target image}. Your task is to determine if the \texttt{text description} accurately describes the transition from the \texttt{source image} to the \texttt{target image}.
\vspace{0.3em}

\textbf{Failure Criteria:} \\
The text description is considered a 'fail' if it meets \textbf{ANY} of the following conditions:
\begin{enumerate}[topsep=2pt, itemsep=2pt, partopsep=0pt, parsep=0pt]
    \item \textbf{Description Mismatch:} The text does not accurately reflect the actual changes between the source and target images. (e.g., The text describes "making the sky blue," but the actual change from source to target shows the sky turning red).
    \item \textbf{Subject Inconsistency:} The core subject or scene in the target image is fundamentally different from the source image, and this difference is \textbf{not} mentioned in the text description. (e.g., The source shows a dog, the target shows a cat, but the text only mentions "removing the background"). However, if the subject basically belongs to the same category, it is acceptable.
    \item \textbf{Transition Gap:} The text fails to describe significant visible changes between the source and target images, leaving important transitions unexplained.
    \item \textbf{Over-Description:} The text describes changes that are not actually present in the transition from source to target image.
\end{enumerate}
\vspace{0.3em}

Here is the text description: [FILL\_THE\_QUERY] 
\vspace{0.3em}

\textbf{Required Output Format:} \\
Please respond strictly in json format, without any additional comments. The json should contain two keys: "verdict" and "reason".

{\ttfamily
\vspace{0.5em}
\noindent
\{ \\
\quad verdict: [pass / fail] \\
\quad reason: [If "fail", provide a brief, specific reason based on the Failure Criteria.] \\
\}
}
\vspace{0.8em}

\textbf{Examples of "fail" Reasons:}
\begin{itemize}[topsep=2pt, itemsep=2pt, partopsep=0pt, parsep=0pt]
    \item "Reason: The text describes changing the car's color to red, but no color change is visible between source and target images."
    \item "Reason: The text fails to mention the significant change in background scenery from urban to rural."
    \item "Reason: The text describes adding a dog, but the target image shows a cat was added instead."
\end{itemize}

\end{tcolorbox}
\caption{Prompt used to assess the match among a source image, a text description, and a target image, serving as a quality-control checkpoint in our data-filtering pipeline.}
\label{fig-image-pair-matching}
\end{figure*}

\end{document}